\documentclass[11pt]{article}

\usepackage[final]{acl}

\usepackage{times}
\usepackage{latexsym}
\usepackage{xcolor}
\usepackage{url}
\usepackage[T1]{fontenc}

\usepackage[utf8]{inputenc}

\usepackage{microtype}
\usepackage{inconsolata}
\usepackage{amsmath}
\usepackage{amssymb}

\usepackage{graphicx}
\usepackage{algorithm}
\usepackage{algorithmic}
\usepackage{xcolor}
\usepackage{pifont}
\usepackage{natbib}
\usepackage{caption}
\usepackage{booktabs}
\usepackage{multirow}
\usepackage{amsfonts}
\usepackage{amsthm}
\usepackage{bm}
\usepackage{tcolorbox}
\usepackage{enumitem}
\usepackage{subcaption}
\usepackage{makecell}
\usepackage{subcaption}
\usepackage[table]{xcolor}
\usepackage{booktabs}
\title{Beyond High-Entropy Exploration: Correctness-Aware Low-Entropy Segment-Based Advantage Shaping for Reasoning LLMs}

\author{
 \textbf{Xinzhu Chen\textsuperscript{1}},
 \textbf{Xuesheng Li\textsuperscript{2}},
  \textbf{Zhongxiang Sun\textsuperscript{3}},
 \textbf{Weijie Yu\textsuperscript{2}\thanks{Corresponding authors.}}
\\
 \textsuperscript{1}Beijing University of Posts and Telecommunications, China\\
 \textsuperscript{2}School of Artificial Intelligence and Data Science,\\ University of International Business and Economics, China\\
 \textsuperscript{3}Gaoling School of Artificial Intelligence, Renmin University of China, China\\
\texttt{\{c1456355244\}@gmail.com}\\
}

\begin{document}
\maketitle
\begin{abstract}
Reinforcement Learning with Verifiable Rewards (RLVR) has become a central approach for improving the reasoning ability of large language models. Recent work studies RLVR through token entropy, arguing that high-entropy tokens drive exploration and should receive stronger updates. However, they overlook the fact that most of a reasoning trajectory consists of low-entropy segments that encode stable and reusable structural patterns. Through qualitative and quantitative analyses, we find that the overlap of low-entropy segments across correct responses strongly correlates with model accuracy, while overlaps involving incorrect responses exhibit stable but unproductive patterns.
Motivated by these findings, we propose LESS, a correctness-aware reinforcement framework that performs fine-grained advantage modulation over low-entropy segments. LESS amplifies segments unique to correct responses, suppresses those unique to incorrect ones, and neutralizes segments shared by both, while preserving high-entropy exploration in the underlying RL algorithm. Instantiated on top of the popular GRPO, LESS consistently improves accuracy over strong RL baselines across three backbones and six math benchmarks, achieves stronger robustness of the performance floor.


\end{abstract}

\section{Introduction}
The reasoning capability of Large Language Models (LLMs) plays a central role in tasks such as mathematics~\citep{guo2025distill,chen2024MathField}, programming~\citep{zhipu2025glm,wei2025SWE-RL,da2025program}, science problem-solving~\citep{baichuan2025M2,Sellergren2025science,jing2025science}, and legal analysis~\citep{legal2,legal1}. Reinforcement Learning with Verifiable Rewards (RLVR) has emerged as an effective approach for improving reasoning reliability, where the correctness of the final answer is used as a reward signal to update the model. 
Representative RLVR methods~\cite{shao2024grpo,yu2025dapo} typically apply policy updates uniformly across all tokens in a generated sequence.

Recent studies have argued that different parts of a reasoning sequence contribute differently to the final outcome, and that RLVR training should take this into account. A set of works examine these differences through the lens of token entropy. They observe that high-entropy tokens often correspond to exploratory reasoning steps, where the model tests alternative solution paths. For example,~\citet{cui2025entropyMechanism} show that training encourages the model to explore uncertain reasoning branches; ~\citet{zhang2025EDGE-GRPO} encourage diversity in correct attempts by adjusting update strength in high-entropy regions; and~\citet{cheng2025Reasoning_with_Exploration}  show that increasing entropy can improve the ability to search for solutions. Most notably, ~\citet{wang2025_8020} study demonstrates that only a small subset of tokens with high entropy disproportionately influence reasoning outcomes, suggesting that RL training should focus attention on these regions.

\begin{figure*}[ht]
  \centering 
  \includegraphics[width=\textwidth]{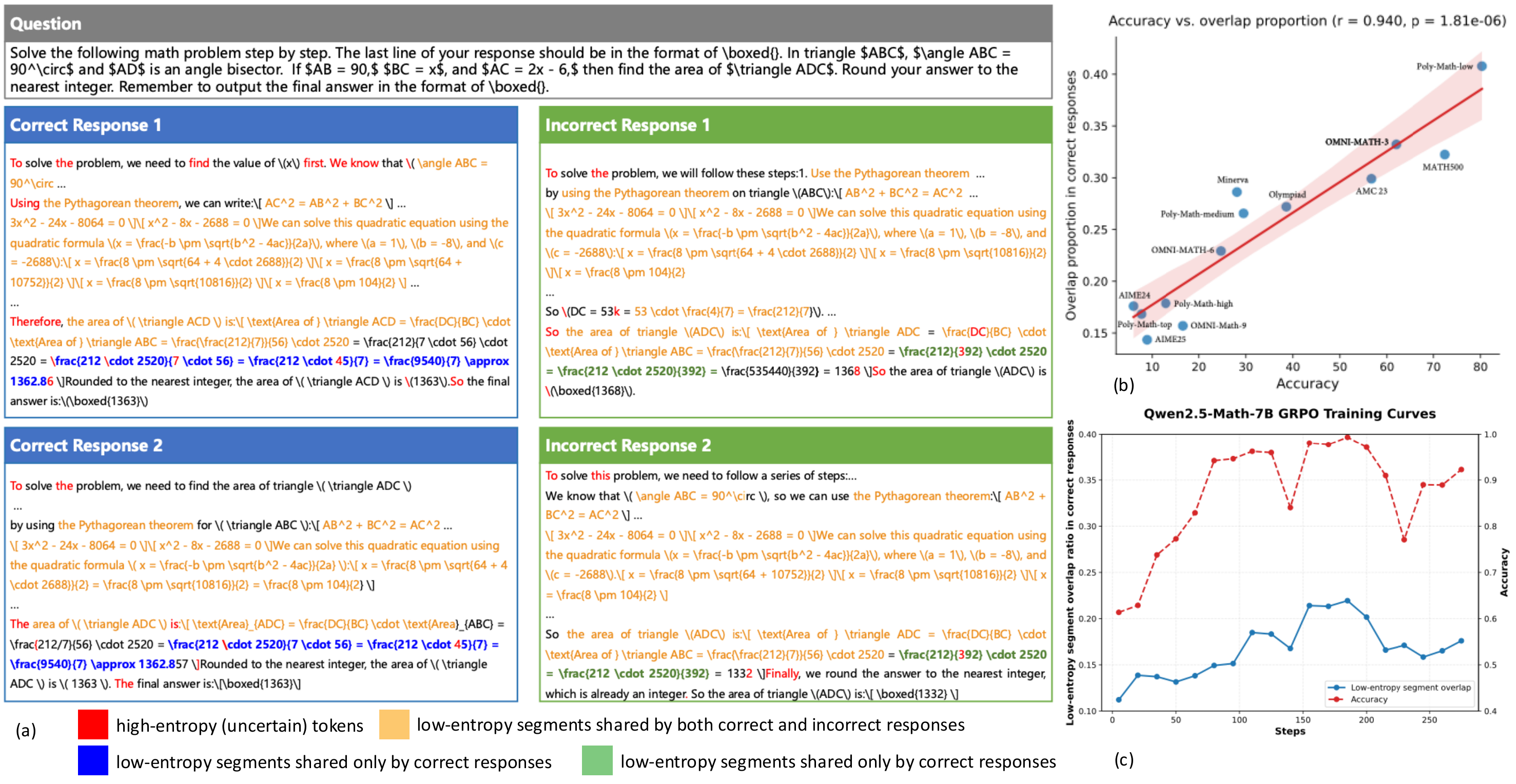}
  \caption {
  Low-entropy analysis reveals stable reasoning behaviors. 
  \textbf{Left}: a case study where correct and incorrect responses exhibit shared and distinct low-entropy segments.
  \textbf{Right-top}: 
  Across math benchmarks, accuracy strongly correlates with low-entropy segment overlap in correct responses (Pearson r, p).
\textbf{Right-bottom}: 
During GRPO training of Qwen2.5-Math-7B, both accuracy and low-entropy overlap rise together, showing that performance gains emerge alongside the stabilization of reasoning patterns.
}
  \label{fig:fig1}
\end{figure*}

While existing entropy-based approaches have shown promising results, they focus almost entirely on high-entropy tokens, treating these points as the main drivers of exploration in reasoning. This overlooks that most of a reasoning sequence is composed of low-entropy segments (e.g.,~\citet{wang2025_8020} 
treat about 80$\%$ of tokens in a response as low-entropy), which form the stable structural scaffold that shapes how the solution is carried out. To examine the role of these low-entropy segments, we conduct both qualitative and quantitative analyses on RLVR-trained models.
\textbf{First}, as shown in Fig~\ref{fig:fig1}(a), correct responses share consistent low-entropy segments that reflect coherent and productive reasoning steps, while incorrect responses also display their own repeated low-entropy patterns that represent stable but unproductive reasoning habits. 
\textbf{Second}, cross-dataset evaluation in Fig~\ref{fig:fig1}(b) demonstrates a strong positive correlation between Qwen2.5-Math-7B accuracy and the overlap of low-entropy segments in correct responses, indicating that this relationship holds across tasks and model setups.
\textbf{Third}, from the training dynamics in Fig~\ref{fig:fig1}(c), we observe that the model accuracy and the overlap of low-entropy segments across correct responses increase together, showing that improvements in reasoning ability are accompanied by the consolidation of shared structural patterns. 
These observations suggest that simply emphasizing high-entropy regions is insufficient, and that the treatment of low-entropy segments is directly related to whether useful or harmful reasoning routines are reinforced. The detailed preliminary analyses are described in §\ref{sec:preliminary}.

Building on these observations, we introduce \textbf{Low-Entropy Segment Shaping (LESS)}, a reinforcement learning with verifiable rewards framework that treats low-entropy structure as an explicit training signal. LESS inserts an entropy-aware segmentation step into the policy update. For each generated solution, it splits the trajectory into high-entropy exploration tokens and contiguous low-entropy segments, and then aggregates how often each segment appears in correct versus incorrect responses within a rollout group. These statistics are used to rescale token-level advantages in a structured way: segments that occur only in correct trajectories receive amplified positive advantages, segments that occur only in incorrect trajectories receive amplified negative advantages, and segments that co-occur in both are neutralized, while high-entropy tokens keep their original RL updates. In this way, LESS strengthens reusable reasoning structure and suppresses repeated failure patterns without harming exploration. The framework is agnostic to the underlying RLVR objective. In this work, we instantiate it on top of Group Relative Policy Optimization (GRPO)~\cite{shao2024grpo}, a widely used multi-sample training method for reasoning LLMs. 
Extensive experiments conducted on six reasoning benchmark demonstrate LESS outperforms popular baseline across almost all tasks and model scales (1.5B, 7B math-tuned, and 7B base). In particular, it yields notable improvements on AIME24/25 and AMC23, where stable multi-step reasoning is essential. Moreover, compared to vanilla GRPO, LESS markedly reduces the worst-case dispersion among sampled responses.
This aligns with the core goal of LESS—to reinforce beneficial structural segments and suppress misleading ones—ultimately producing more stable and reliable policy updates.

In summary, our contributions are three-fold: (1) We introduce a segment-level perspective on RLVR that distingsuishes low-entropy segments by correctness, revealing stable structural patterns in LLM reasoning. (2) We propose LESS, a plug-in algorithm that reweights token-level advantages by segment statistics, amplifying low-entropy segments unique to correct trajectories, suppressing those unique to incorrect ones, and neutralizing shared segments. (3) We show that LESS consistently improves accuracy across six mathematical reasoning benchmarks and three backbones, while also improving robustness under worst@K and reducing variance across sampled rollouts.

\section{Preliminary Analysis}
\label{sec:preliminary}
We examine low-entropy segments as an indicator of stable reasoning behavior in LLMs and study how these signals relate to model correctness and performance.

We begin by conducting a qualitative experiment to visualize the entropy structure of multiple responses produced for the same question. Specifically, we analyze the token-level entropy patterns of Qwen2.5-Math-7B on the mathematical reasoning dataset~\citep{hendrycks2021Math}, which enables us to separate responses into stable and unstable regions and to identify the parts of the model’s reasoning that remain consistently preserved across different outputs. As shown in Fig.~\ref{fig:fig1} (left), high-entropy tokens (in \textcolor{red}{red}) mark unstable regions where the model varies its reasoning, while low-entropy segments reveal stable structures that the model consistently reuses. Within these low-entropy segments, we observe three distinct patterns. (1) Segments shared only by correct responses (in \textcolor{blue}{blue}) correspond to productive reasoning steps that reliably support the correct solution. (2) Segments shared only by incorrect responses (in \textcolor{green}{green}) reflect stable but unproductive reasoning habits, for example, the repeated computation ``$\frac{212}{392}\cdot 2520=\frac{212 \cdot 2520}{392} = ?$''. (3) Segments shared by both correct and incorrect responses (in \textcolor{orange}{orange}) capture general reasoning components that are stable but not predictive of correctness—for instance, invoking ``the Pythagorean theorem,'' which provides a common derivation framework but is not the source of the subsequent correct or incorrect calculations. This 
evidence shows low-entropy segments encode structured reasoning behaviors that differentiate effective and ineffective model responses.

To test whether the qualitative patterns extend beyond a single example, we measure the overlap of low-entropy segments across correct responses for a range of math benchmarks, including AIME24, AIME25, AMC23, MATH500, Minerva, and the Omni-MATH series. As shown in Fig.\ref{fig:fig1} (right-top), benchmark accuracy is strongly correlated with the degree of low-entropy overlap across correct responses (Pearson $r=0.94$ and  $p\text{-value}=1.81e^{-6}$). Benchmarks with higher accuracy, such as MATH500 and Omni-MATH-3, exhibit clear clustering toward higher overlap ratios, while lower-accuracy benchmarks show weaker consistency in their stable segments. The fitted regression line further highlights this trend, indicating that stronger task performance is associated with more consolidated reasoning structure and greater reuse of stable low-entropy patterns. Similar positive correlations (Appendix~\ref{app:extra-corr}) are observed across several other backbones.

We further examine how these patterns evolve during learning. Using GRPO training of Qwen2.5-Math-7B, we track both accuracy and low-entropy segment overlap over training steps. As shown in Fig.~~\ref{fig:fig1} (right-bottom), the two trajectories rise together throughout training: early stages display low accuracy and fragmented low-entropy structure, while later stages show increasing stability in low-entropy segments alongside improved accuracy. This synchronous growth suggests that the model’s reasoning becomes more consistent as training progresses and that stable low-entropy segments emerge as the model acquires more reliable reasoning routines. These results confirm that low-entropy overlap reflects not only final performance but also the developmental trajectory of the model’s reasoning behavior. We observe the same co-evolution pattern (Appendix~\ref{app:curves})
on Qwen2.5-Math-1.5B and Qwen2.5-Base-7B.

These results
show that low-entropy segments provide a reliable signal for understanding and guiding model reasoning. They capture stable computational routines that distinguish correct from incorrect behavior, reflect the degree of structural consistency across benchmarks, and track the development of reasoning stability during learning. These observations suggest low-entropy segments can serve as informative targets for optimization, enabling the model to strengthen productive reasoning routines while suppressing unproductive ones.

\section{Methodology}
Motivated by these findings, we propose Low-Entropy Segment Shaping (LESS), an RLVR framework that improves reasoning stability by reshaping token-level advantages using statistics of low-entropy segments across sampled responses. LESS is compatible with standard RLVR algorithms and can be used as a plug-in module, and we instantiate it on top of GRPO~\citep{shao2024grpo} in this work.

\subsection{LESS: Low-Entropy Segment Shaping}
Given an input question $q$, the policy generates a group of responses 
$\mathcal{G}=\{O_1,\ldots,O_G\}$,
LESS detects low-entropy segments and shapes corresponding advantages as follows:

\paragraph{Entropy-based segment extraction.}
For a 
$O_i=[t_1,\ldots,t_{|O_i|}]$,
the entropy of token $t_j$ is
\begin{equation}
\small
\mathcal{H}_{t_j}
=-
\!\sum_{v\in V}
\pi_{\theta_\text{old}}(v\mid x,O_i{<}j)
\log\pi_{\theta_\text{old}}(v\mid x,O_i{<}j).
\end{equation}
Following \citet{wang2025_8020}, we compute an entropy threshold $\tau_i$ for each response $O_i$ as the $h$-quantile of its token entropies $\mathcal{H}_{t_j}$. We then treat high-entropy tokens as isolated positions and group consecutive low-entropy tokens into contiguous spans. A minimum length $\mu$ is used to filter out trivial low-entropy spans (such as punctuation or very short frequent phrases). This gives three types of entropy-based structures:
\begin{equation}
\small
\begin{aligned}
\mathcal{S}_i^{\text{high}} &= 
\{t_j \in O_i \mid \mathcal{H}_{t_j} \ge \tau_i\},\\
\mathcal{S}_i^{\text{frag}} &= 
\{O_i[a{:}b] \mid b-a+1 < \mu,~\forall t_j\!\in[a,b]:~\mathcal{H}_{t_j} < \tau_i\},\\
\mathcal{S}_i^{\text{seg}}  &= 
\{O_i[a{:}b] \mid b-a+1 \ge \mu,~\forall t_j\!\in[a,b]:~\mathcal{H}_{t_j} < \tau_i\},
\end{aligned}
\label{eq:segment-sets}
\end{equation}
where $\mathcal{S}_i^{\text{high}}$ collects individual high-entropy tokens, 
$\mathcal{S}_i^{\text{frag}}$ contains short low-entropy fragments that are likely uninformative, 
and $\mathcal{S}_i^{\text{seg}}$ contains longer low-entropy segments that we regard as structured reasoning candidates.
Then, we aggregate how often each $\mathcal{S}_i^{\text{seg}}$ appears in correct versus incorrect responses within a rollout group.
Let $N_r$ and $N_w$ denote the number of correct and incorrect responses in $\mathcal{G}$.  
For a low-entropy segment $\sigma$, we count its frequency over the group:
\begin{equation}
\small
\begin{aligned}
n_r(\sigma)&=|\{i\mid correct_i=1\land\sigma\in\mathcal{S}_i^{\mathrm{seg}}\}|,\\
n_w(\sigma)&=|\{i\mid correct_i=0\land\sigma\in\mathcal{S}_i^{\mathrm{seg}}\}|.
\end{aligned}
\label{eq:count}
\end{equation}

\paragraph{Advantage shaping.}
LESS modifies the advantage assigned to each token $t_j$ in $O_i$ as:
\begin{equation}
\small
\hat{A}_j^i =
\begin{cases}
A_i,
& t_j\in\mathcal{S}_i^{\mathrm{high}},
\\[3pt]
A_i/N_r,
& t_j\in\mathcal{S}_i^{\mathrm{frag}},~correct_i=1,
\\[3pt]
A_i/N_w,
& t_j\in\mathcal{S}_i^{\mathrm{frag}},~correct_i=0,
\\[3pt]
0,
& \sigma_j^i\!\in\!\mathcal{S}_i^{\mathrm{seg}},~n_r>0,~n_w>0,
\\[3pt]
(n_r/N_r)A_i,
& \sigma_j^i\!\in\!\mathcal{S}_i^{\mathrm{seg}},~n_r>0,~n_w=0,
\\[3pt]
(n_w/N_w)A_i,
& \sigma_j^i\!\in\!\mathcal{S}_i^{\mathrm{seg}},~n_r=0,~n_w>0.
\end{cases}
\label{eq:shaping}
\end{equation}
This rule:
(i) preserves exploratory high-entropy behavior,
(ii) reinforces stable segments unique to correct responses,
(iii) penalizes those unique to incorrect responses,
(iv) ignores ambiguous segments shared by both groups.
The full pseudocode of LESS and its time complexity analysis are given in Appendix~\ref{app:less-algorithm}. 

\subsection{Instantiating LESS with GRPO}
LESS is designed as a generic advantage–shaping framework and can be applied to current RLVR methods that computes token- or sequence-level advantages.
In this work, we instantiate LESS using the GRPO~\citep{shao2024grpo}, which has become the standard multi-sample
training paradigm for mathematical and logical reasoning models.  
GRPO is particularly suitable for our setting because it (i) generates a group of
responses for each query, allowing entropy-based statistics to be computed across
samples, and (ii) performs stable clipped-ratio updates that interact well with our
advantage shaping.
Specifically, for an input query $q$, the policy produces $G$ responses with rewards
\{$r_1,\ldots,r_G$\}.  
GRPO standardizes these rewards to obtain group-relative advantages:
\begin{equation}
\small
A_i=\frac{r_i-\mathrm{mean}(r_{1:G})}{\mathrm{std}(r_{1:G})}.
\label{eq:grpo-adv}
\end{equation}
GRPO then updates the policy by maximizing a clipped likelihood-ratio objective
regularized by a KL constraint toward a reference policy:
{
\small
\begin{align}
J_{\mathrm{GRPO}}(\theta)
&=
\mathbb{E}\!\left[
   \frac{1}{G}\sum_{i=1}^{G}
      \frac{1}{|o_i|}
      \left(
         \min(\alpha_i A_i,\; \tilde{\alpha}_i A_i)
         - \kappa_i
      \right)
\right], \notag\\[4pt]
\text{where}~~\alpha_i 
&= 
\frac{\pi_{\theta}(o_i \mid x)}
     {\pi_{\theta_{\mathrm{old}}}(o_i \mid x)}, 
~~
\tilde{\alpha}_i
= 
\operatorname{clip}(\alpha_i, 1-\epsilon, 1+\epsilon), \notag\\[4pt]
\kappa_i
&= 
\beta\, D_{\mathrm{KL}}\!\left(
\pi_{\theta}\,\|\,\pi_{\mathrm{ref}}
\right),
\label{eq:grpo_obj}
\end{align}
}
GRPO’s group-wise credit assignment is well aligned with LESS, since the same
group of responses used for reward normalization is also used by LESS to compute
low-entropy statistics.  
Replacing $A_i$ in Eq.~\ref{eq:grpo_obj} with the shaped advantage
$\hat{A}_j^i$ in Eq.~\ref{eq:shaping} yields our LESS-GRPO training objective.

\section{Experiments}
We answer the following research questions with experiments: \textbf{RQ1}: How does LESS perform across diverse benchmarks compared with strong RL baselines?
\textbf{RQ2}: How does LESS influence the training dynamics of LLM reasoning, compared with GRPO, in terms of accuracy growth, stability.
\textbf{RQ3}: How do LESS and GRPO differ in the evolution of entropy-based reasoning structures during training?
\textbf{RQ4}: Does LESS improve worst-case reasoning robustness compared with GRPO across different model sizes?
\textbf{RQ5}: How sensitive is LESS to the minimum segment-length $\mu$, and how does varying $\mu$ affect the stability and final accuracy of reinforcement-learning-based reasoning?

\subsection{Experimental Setup}
\paragraph{Datasets and evaluation metrics.} Following~\citep{Shen2025Aent}, we train the models on the MATH dataset~\citep{hendrycks2021Math}, which contains 7,500 problems spanning algebra, geometry, counting, probability, number theory, and other areas. The dataset is widely adopted in LLM reasoning research due to its breadth and the step-wise reasoning it elicits, making it particularly suitable for entropy-based structure analysis. 

In terms of the evaluation, we assess the trained models on a suite of standard mathematical reasoning benchmarks:
MATH500~\citep{hendrycks2021Math},
Minerva Math~\citep{lew2022Minerva},
OlympiadBench~\citep{He2024Olympiad},
AIME’24, and AIME’25~\citep{li2024AIME}.
These datasets collectively cover varying difficulty levels and reasoning types, allowing us to examine whether LESS consistently improves reasoning stability.
For all benchmarks except AIME, we report accuracy under greedy decoding, which is commonly used in math reasoning evaluation.
For AIME’24/25, we follow prior works\citep{yu2025dapo,zheng2025FR3E,yue2025vapo} and compute the avg@32 accuracy by averaging predictions over 32 sampled rollouts. This protocol reduces the variance introduced by integer-answer formats and ensures fair comparison across RL-trained models.

\paragraph{Backbone LLM and baselines.}  We evaluate LESS on three Qwen2.5 variants: Qwen2.5-Math-1.5B, Qwen2.5-Math-7B, and the general-purpose Qwen2.5-7B model~\citep{qwen2024qwen}.
These models allow us to test LESS across (i) different parameter scales and (ii) models with and without domain-specific pretraining. This diversity is essential for assessing whether the proposed token-level shaping mechanism generalizes beyond math-specialized checkpoints.

We compare LESS against strong RLVR systems including: \textbf{GRPO}~\citep{shao2024grpo}, the canonical multi-sample policy-gradient method and the underlying backbone of many reasoning RL pipelines;
\textbf{Forking Tokens}~\citep{wang2025_8020}, an approach that identifies repeated reasoning fragments to adjust token-level credit assignment.
\textbf{KL-Cov}~\citep{cui2025entropyMechanism}, an entropy-based mechanism that modulates KL penalties using covariance between reward and token log-probs.
These baselines represent the closest lines of work involving token-level structure, multi-sample variance reduction, and entropy-informed regularization, making them the most relevant comparisons for evaluating LESS.

\paragraph{Implementation details.} We conduct experiments using the VeRL \citep{sheng2024Verl} framework for reinforcement learning with LLMs on 8 NVIDIA A100-40G GPUs. The training setup includes a batch size of 512, a learning rate of $1 \times 10^{-6}$, and a clip range between 0.2 and 0.28. Each response sequence is up to 3k tokens in length. The mini-batch size is set to 32. The temperature is 1.0 for training and 0.1 for evaluation. 
Following prior RLVR work~\citep{liu2025GHPO,yang2025SSPO}, we perform 8 rollouts per promp and do not use entropy regularization or KL penalties during training (KL coefficient = 0, entropy loss = 0), allowing us to isolate the effect of LESS from other types of regularization. This choice also reveals whether LESS alone can stabilize reasoning trajectories without relying on heavy KL anchoring. We use a unified prompt format for all experiments, and the exact template is provided in Appendix~\ref{app:prompt-template}.

\begin{table*}[t]
\centering
\setlength{\tabcolsep}{4pt}
\small
\caption{Overall performance on mathematical reasoning benchmarks.
AIME24/25 are evaluated with avg@32; other benchmarks use @avg1.
Bold numbers are the best in each column.
`$\dagger$' indicates the model significantly outperforms all baseline models with paired t-tests at $p < 0.05$ level.}
\label{tab:overall}
\begin{tabular}{lccccccc}
\toprule
Method & AIME24 & AIME25 & AMC23 & MATH500 & Minerva & Olympiad & Avg \\
\midrule
\multicolumn{8}{c}{\textbf{Qwen2.5-Math-1.5B}}\\
\midrule
Base LLM   &  6.2 &  3.8 & 37.5 & 58.6 & 15.8 & 26.5 & 24.7 \\
GRPO   & 21.6 &  6.6 & 57.5 & 74.8 & 27.2 & \textbf{39.6} & 37.8 \\
Forking Tokens  & 21.6 &  7.0 & 60.0 & 75.6 & 29.0 & 38.5 & 38.6 \\
KL-Cov  & 22.0 &  9.2 & 57.5 & \textbf{75.8} & 27.9 & 38.4 & 38.4 \\
\rowcolor{gray!10}
LESS (ours) & \textbf{26.2}$^\dagger$ & \textbf{12.4}$^\dagger$ & \textbf{62.5}$^\dagger$ & 75.2 & \textbf{30.8}$^\dagger$ & \textbf{39.6} & \textbf{41.1}$^\dagger$ \\
\midrule
\multicolumn{8}{c}{\textbf{Qwen2.5-Math-7B}}\\
\midrule
Base LLM   &  6.0 &  8.9 & 57.5 & 58.6 & 28.7 & 38.0 & 32.9 \\
GRPO   & 33.3 & 13.8 & 65.0 & 79.8 & \textbf{38.6} & 44.9 & 45.9 \\
Forking Tokens   & \textbf{36.6} & 14.2 & 67.5 & 78.2 & 37.8 & 42.2 & 46.0 \\
KL-Cov  & 35.2 & 14.1 & \textbf{70.0} & 78.6 & 38.2 & 43.7 & 46.6 \\
\rowcolor{gray!10}
LESS (ours) & 36.0 & \textbf{15.6}$^\dagger$ & \textbf{70.0} & \textbf{81.6}$^\dagger$ & 37.8 & \textbf{45.7}$^\dagger$ & \textbf{47.7}$^\dagger$ \\
\midrule
\multicolumn{8}{c}{\textbf{Qwen2.5-Base-7B}}\\
\midrule
Base LLM    &  3.2 &  5.2 & 37.5 & 53.4 & 18.0 & 23.9 & 23.5 \\
GRPO   & 18.9 & 11.9 & 60.0 & 76.8 & 37.1 & 40.9 & 40.9 \\
Forking Tokens  & 20.2 & 12.3 & 62.5 & 77.8 & 37.5 & 40.6 & 41.8 \\
KL-Cov  & 18.7 & 10.8 & 60.0 & 77.0 & \textbf{37.9} & 40.8 & 40.9 \\
\rowcolor{gray!10}
LESS (ours) & \textbf{20.5}$^\dagger$ & \textbf{13.1}$^\dagger$ & \textbf{67.5}$^\dagger$ & \textbf{78.6}$^\dagger$ & 37.1 & 40.8 & \textbf{42.9}$^\dagger$ \\
\bottomrule
\end{tabular}
\label{tab:overall}
\end{table*}

\begin{figure*}[ht]
  \centering 
  \includegraphics[width=\textwidth]{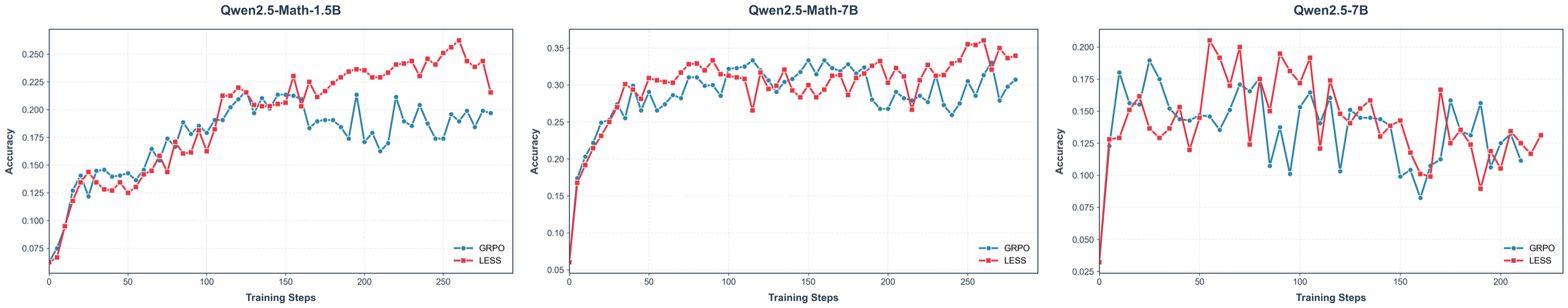}
  \caption {Training dynamics (accuracy over training) of GRPO and LESS across three backbones.}
  \label{acc}
\end{figure*}

\subsection{Overall Performance}
To answer \textbf{RQ1}, we conduct extensive experiments on seven reasoning benchmark datasets. Table~\ref{tab:overall} shows that LESS consistently outperforms all RL baselines across the three Qwen2.5 backbones. 
On the math-specialized 1.5B model, LESS achieves the highest average score, with clear gains on challenging tasks such as AIME24, AIME25, and AMC23, indicating that entropy-guided advantage shaping improves reliability over token-level clipping and forking-based updates.
For the stronger 7B math model, LESS further improves the average to 47.7, with the largest margins appearing on MATH500 and OlympiadBench, suggesting that the method strengthens multi-step reasoning structures rather than solving isolated symbolic steps.
Finally, on the 7B base model without math specialization, LESS still provides consistent gains, showing that the approach generalizes beyond math-aligned checkpoints.
Overall, LESS delivers the best average performance in every setting, demonstrating robust improvements across both easy and hard reasoning benchmarks.

\begin{figure*}[ht]
  \centering 
  \includegraphics[width=\textwidth]{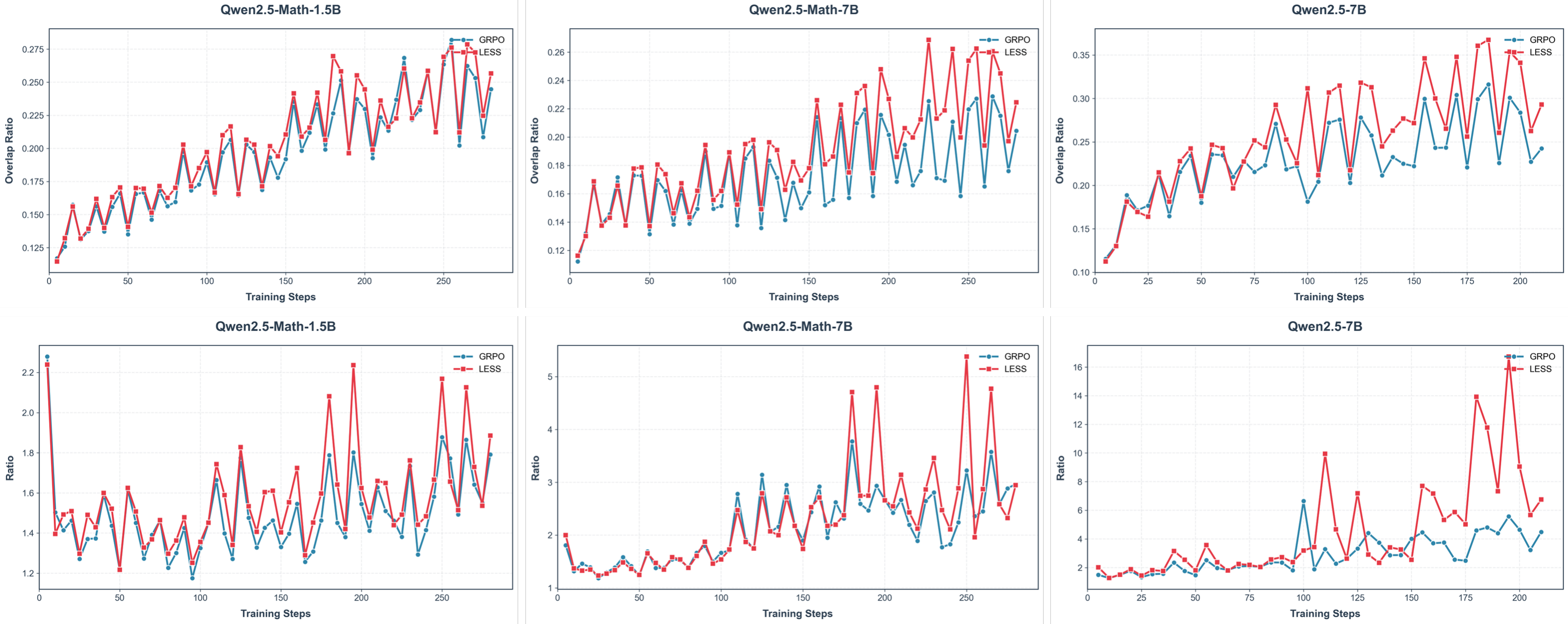}
  \caption {Training-dynamics comparison between LESS and GRPO across three model sizes. \textbf{Top}: Ratio of low-entropy segments that overlap exclusively among correct responses (higher is better). \textbf{Bottom}: Ratio between the entropy of incorrect responses and correct responses (higher indicates that incorrect answers remain exploratory). LESS consistently strengthens productive low-entropy structures while preventing premature entropy collapse in incorrect trajectories.}
  \label{ratio_of_overlap_and_entropy}
\end{figure*}

\subsection{Training Dynamics Analysis}
To answer \textbf{RQ2}, we examine LESS and GRPO on how accuracy and entropy-based structure evolve throughout training.
As shown in Fig.~\ref{acc}, across all three backbones, LESS exhibits a characteristic two-phase learning pattern. In the early stage, LESS improves slightly slower than GRPO because its advantage shaping reduces the update magnitude on low-entropy segments until the model accumulates enough evidence to distinguish productive from unproductive ones. However, as training progresses, LESS consistently surpasses GRPO and maintains a higher accuracy plateau, indicating more stable policy improvement.

To answer \textbf{RQ3}, we analyze how the overlap of correct low-entropy segments and the entropy structure of incorrect responses evolve during reinforcement learning.
the top row of Fig.~\ref{ratio_of_overlap_and_entropy} shows that LESS consistently yields a higher overlap ratio of correct-only low-entropy segments as training progresses. This indicates that LESS explicitly amplifies structurally productive reasoning patterns that appear repeatedly in correct trajectories. In contrast, GRPO shows a flatter or noisier trend, suggesting that it does not reliably consolidate these stable reasoning components. The clearer upward trajectory of LESS reveals that the model is progressively internalizing reusable, high-quality reasoning routines rather than relying on isolated or brittle solution paths.

The bottom row of Fig.~\ref{ratio_of_overlap_and_entropy} further shows that LESS maintains a higher entropy ratio between incorrect and correct responses, meaning that incorrect trajectories remain more uncertain. This separation is desirable, that is, LESS avoids prematurely stabilizing low-entropy segments that consistently lead to wrong answers, thereby reducing the risk of ``locking in” systematic errors. GRPO, however, frequently collapses the entropy gap, causing incorrect responses to become low-entropy as well—an indication that harmful patterns are becoming entrenched in the policy.

These results show that LESS not only improves performance, but also progressively increases the overlap of correct low-entropy segments while keeping incorrect trajectories uncertain, creating a clear structural separation between productive and unproductive reasoning.

\begin{table}[t]
\caption{Worst-case reasoning performance (worst@$k$) across three backbones. For each prompt, the worst-performing sample among $k$ rollouts is selected and averaged over the dataset. LESS consistently improves worst-case accuracy across all settings.}
  \centering
  \small
  \begin{tabular}{lccc}
    \toprule
    \textbf{Method} & \textbf{worst@32}  & \textbf{worst@16}  & \textbf{worst@8}\\
    \midrule
    \multicolumn{4}{c}{\textit{Qwen2.5-math-1.5b}} \\
    \midrule
    GRPO    &6.8      &8.0     &10.8 \\
    LESS     &\textbf{12.9}      &\textbf{15.6}    &\textbf{18.2}  \\
    \midrule
    \multicolumn{4}{c}{\textit{Qwen2.5-math-7b}} \\
    \midrule
    GRPO    & 13.4     &17.4     &20.6 \\
    LESS     &\textbf{21.2}     &\textbf{22.2}    &\textbf{24.3}  \\
    \midrule
    \multicolumn{4}{c}{\textit{Qwen2.5-7b}} \\
    \midrule
    GRPO    & 10.3     &11.3     &12.9 \\
    LESS     & \textbf{11.1}     &\textbf{12.0}    &\textbf{13.4}   \\
    \bottomrule
  \end{tabular}
  
  \label{worst@k}
\end{table}
\begin{figure}[ht]
\centering
  \includegraphics[width=0.8\columnwidth]{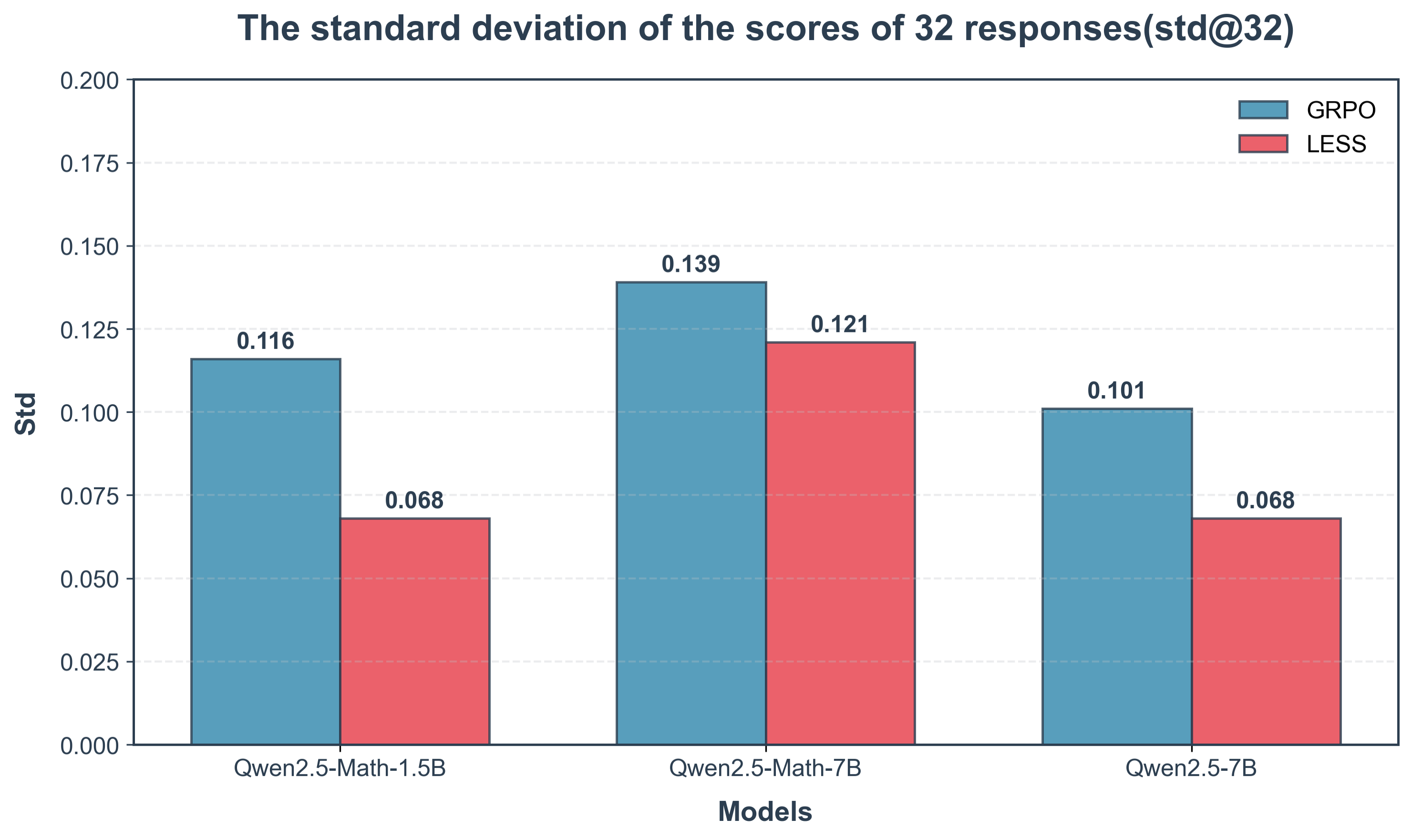}
  \caption {Standard deviation of 32 sampled responses (std@32). LESS reduces response-level variability across all backbones, indicating more stable and less volatile reasoning behavior compared with GRPO.}
  \label{std}
\end{figure}
\subsection{Robustness Under Worst-Case Sampling}
To answer \textbf{RQ4} and examine the robustness of the learned policy, we employ the worst@$k$ metric, which selects the lowest-scoring output among $k$ sampled responses and averages this worst-case score across prompts. This metric directly measures how the model behaves in its most vulnerable failure modes.

As shown in Table.~\ref{worst@k}, across all three backbones, LESS consistently improves worst-case accuracy. For Qwen2.5-Math-1.5B and Qwen2.5-Math-7B, LESS achieves substantial gains, raising worst@32 by +6.1 and +7.8 points respectively, with positive margins maintained as $k$ decreases. This pattern shows that LESS not only lifts average performance but also strengthens the weakest trajectories, suppressing brittle low-entropy patterns that GRPO tends to reinforce. Even on the non-math Qwen2.5-7B model, LESS produces steady improvements, indicating that its robustness effects generalize beyond specialized mathematical checkpoints.

The variance results shown in Fig.~\ref{std} further reinforce this finding: LESS consistently reduces the standard deviation of sampled rollouts (std@32), yielding more stable and predictable reasoning behavior. Together, the worst@$k$ and variance metrics demonstrate that LESS raises the floor of model performance while simultaneously mitigating response-level volatility.

These results show that LESS meaningfully improves robustness by raising the floor of model performance while simultaneously reducing instability across sampled rollouts.

\subsection{Impact of the Segment Length}
\begin{figure}[t]
\centering
  \includegraphics[width=0.8\linewidth]{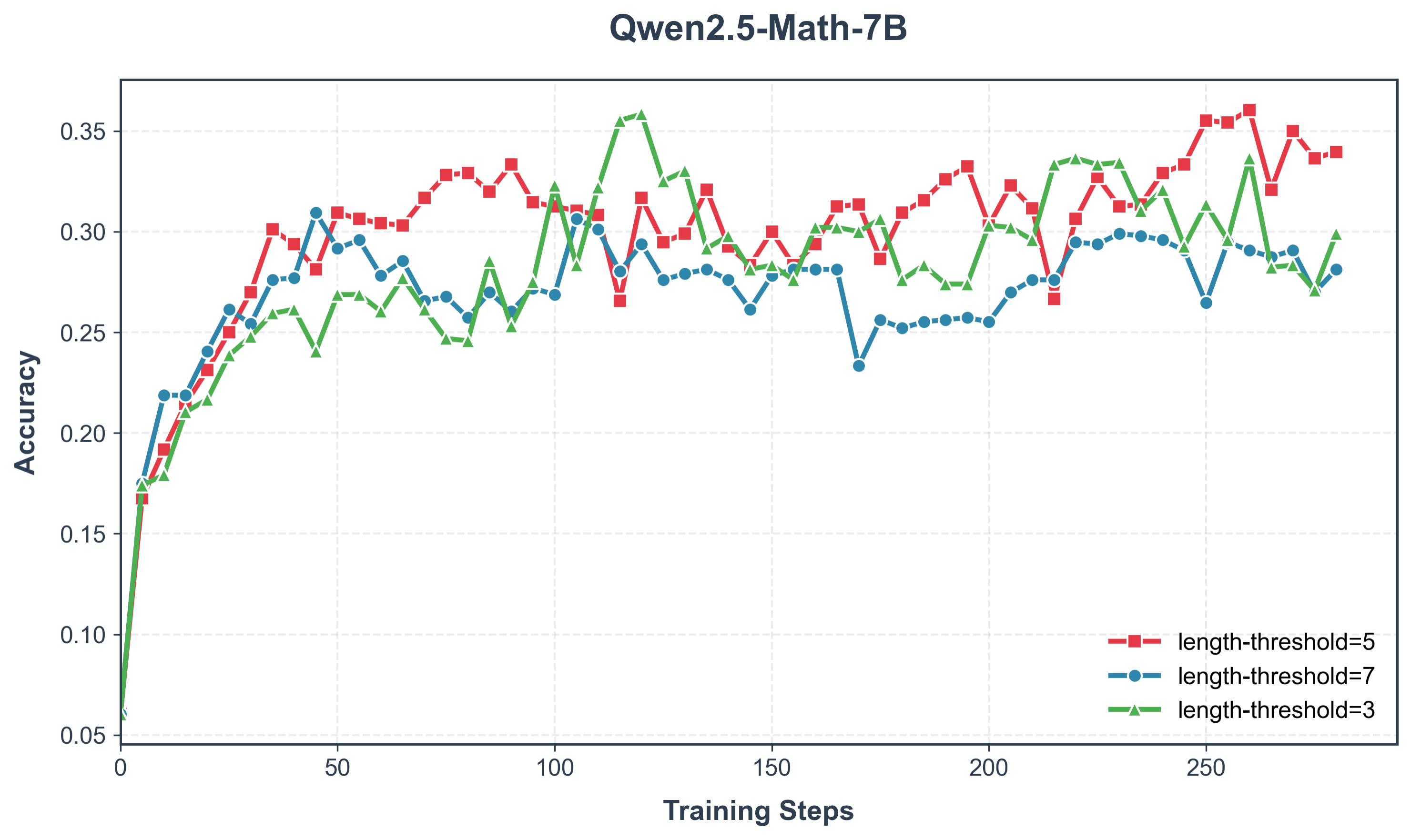}
  \caption {Effect of the low-entropy segment-length threshold $\mu$ on training dynamics for Qwen2.5-Math-7B. We compare $\mu=3,5,7$ and report accuracy over training steps.}
  \label{fig:mu}
\end{figure}
To answer \textbf{RQ5}, we study the sensitivity of LESS to the minimum segment-length threshold $\mu$, we vary $\mu \in \{3,5,7\}$ and track training dynamics on Qwen2.5-Math-7B.
As shown in Figure~\ref{fig:mu}, $\mu=5$ produces the most stable and highest final accuracy across the entire training trajectory. A very small threshold ($\mu=3$) makes the model overly sensitive to short, noisy low-entropy fragments, causing the policy to reinforce many spurious local patterns and resulting in pronounced fluctuations. Conversely, a larger threshold ($\mu=7$) filters out too many low-entropy segments, delaying the discovery of reliable reasoning motifs and slowing convergence.

The superior performance of $\mu=5$ suggests that effective low-entropy guidance requires a balance: segments must be long enough to encode meaningful reasoning structure, yet short enough to capture fine-grained patterns that recur across correct trajectories. This indicates that LESS benefits from moderately sized structural units and is robust to reasonable choices of $\mu$, but extremely small or large thresholds degrade the quality of structural signals made available to the policy.

\section{Related Work}
Recent RLVR work has moved beyond scalar rewards and studied finer training signals that shape how credit is assigned inside reasoning traces.
\paragraph{Token credit assignment.}\citet{vassoyan} identify critical tokens in chain-of-thought solutions, that is, decision points where the model is likely to fail, and increase exploration around these tokens by adjusting the KL penalty.
\citet{lincritical} likewise locate tokens that strongly influence incorrect outcomes and show that editing or replacing these tokens can change the final decision.
Other work~\citep{chan2024denseReward,xie2025capo,guo2025SPO,wei2025CreaditAssign} addresses the coarse-grained nature of standard RL feedback by constructing dense, token-level rewards to resolve the credit assignment problem.
These methods demonstrate that tokens within a trajectory should not be treated uniformly, but they still operate on local positions and do not capture how stable patterns repeat across multiple rollouts of the same question.
\paragraph{Entropy-based RL signals.}
\citet{wang2025_8020} split trajectories at high-entropy tokens and update only a subset of tokens, aiming to reduce over-optimization on already confident regions.
\citet{cui2025entropyMechanism} further modulate the KL penalty based on token-level uncertainty, encouraging updates where the model is less certain and damping updates on low-entropy tokens.
In these approaches, high-entropy tokens serve as a proxy for exploration, while low-entropy regions are treated as parts of the trajectory that should be protected from change.
However, they do not distinguish low-entropy patterns that are consistently correct from those that encode repeated mistakes.

\section{Conclusion}
This paper presents a new perspective on training reasoning LLMs: reasoning should be guided at the level of low-entropy segments rather than individual tokens. Building on this insight, we propose LESS, a plug-and-play advantage-shaping framework that selectively amplifies reliable low-entropy reasoning segments and suppresses error-prone ones. Instantiated with GRPO, LESS improves accuracy, stability, and robustness across multiple backbones and benchmarks. The framework consistently strengthens correct reasoning routines, preserves exploration on incorrect trajectories, and raises the worst-case performance of sampled rollouts. These results show that low-entropy structural signals offer a principled and effective handle for guiding RL training of reasoning models.

\section{Limitations}
We believe there is still room for improvement in our work.
Our preliminary analysis and all main experiments are conducted on the Qwen2.5 family. We do not test LESS on other backbones such as Llama, so it is unclear whether the same entropy patterns and gains will hold more broadly. 
In addition, following \citet{wang2025_8020}, we fix the entropy quantile and length threshold to extract low-entropy segments. We do not yet study how different quantiles, adaptive thresholds, or alternative segmentation rules would affect the learned segments and the final performance.
\bibliography{anthology,custom}

\begin{thebibliography}{34}
\providecommand{\natexlab}[1]{#1}

\bibitem[{{5 Team} et~al.(2025){5 Team}, {Zeng}, {Lv}, {Zheng}, {Hou}, {Chen}, {Xie}, {Wang}, {Yin}, {Zeng}, {Zhang}, {Wang}, {Zhong}, {Liu}, {Lu}, {Cao}, {Zhang}, {Huang}, {Wei}, {Cheng}, {An}, {Niu}, {Wen}, {Bai}, {Du}, {Wang}, {Zhu}, {Zhang}, {Wen}, {Wu}, {Xu}, {Huang}, {Zhao}, {Cai}, {Yu}, {Li}, {Ge}, {Huang}, {Zhang}, {Xu}, {Zhu}, {Li}, {Yin}, {Lin}, {Yang}, {Jiang}, {Ai}, {Zhu}, {Wang}, {Pan}, {Wang}, {Sun}, {Li}, {Li}, {Hu}, {Zhang}, {Peng}, {Tai}, {Zhang}, {Wang}, {Yang}, {Liu}, {Zhao}, {Liu}, {Yan}, {Liu}, {Chen}, {Li}, {Zhao}, {Ren}, {Jiao}, {Zhao}, {Yan}, {Wang}, {Gui}, {Zhao}, {Liu}, {Li}, {Li}, {Lu}, {Wang}, {Yuan}, {Li}, {Du}, {Du}, {Liu}, {Zhi}, {Gao}, {Wang}, {Yang}, {Xu}, {Fan}, {Wu}, {Ding}, {Wang}, {Zhang}, {Li}, {Xu}, {Zhao}, {Zhai}, {Du}, {Dong}, {Lei}, {Tu}, {Yang}, {Lu}, {Li}, {Li}, {Shuang-Li}, {Yang}, {Yi}, {Yu}, {Tian}, {Wang}, {Yu}, {Tam}, {Liang}, {Liu}, {Wang}, {Jia}, {Gu}, {Ling}, {Wang}, {Fan}, {Pan}, {Zhang}, {Zhang}, {Fu}, {Zhang}, {Xu}, {Wu}, {Lu}, {Wang}, {Zhou}, {Pan},
  {Zhang}, {Wang}, {Li}, {Su}, {Geng}, {Zhu}, {Yang}, {Li}, {Wu}, {Li}, {Liu}, {Wang}, {Li}, {Zhang}, {Liu}, {Yang}, {Zhou}, {Qiao}, {Feng}, {Liu}, {Zhang}, {Wang}, {Yao}, {Wang}, {Liu}, {Chai}, {Li}, {Zhao}, {Chen}, {Zhai}, {Xu}, {Huang}, {Wang}, {Li}, {Dong}, and {Tang}}]{zhipu2025glm}
{5 Team}, Aohan {Zeng}, Xin {Lv}, Qinkai {Zheng}, Zhenyu {Hou}, Bin {Chen}, Chengxing {Xie}, Cunxiang {Wang}, Da~{Yin}, Hao {Zeng}, Jiajie {Zhang}, Kedong {Wang}, Lucen {Zhong}, Mingdao {Liu}, Rui {Lu}, Shulin {Cao}, Xiaohan {Zhang}, Xuancheng {Huang}, Yao {Wei}, and 152 others. 2025.
\newblock \href {https://doi.org/10.48550/arXiv.2508.06471} {{GLM-4.5: Agentic, Reasoning, and Coding (ARC) Foundation Models}}.
\newblock \emph{arXiv e-prints}, arXiv:2508.06471.

\bibitem[{{Chan} et~al.(2024){Chan}, {Sun}, {Holt}, and {van der Schaar}}]{chan2024denseReward}
Alex~J. {Chan}, Hao {Sun}, Samuel {Holt}, and Mihaela {van der Schaar}. 2024.
\newblock \href {https://doi.org/10.48550/arXiv.2402.00782} {{Dense Reward for Free in Reinforcement Learning from Human Feedback}}.
\newblock \emph{arXiv e-prints}, arXiv:2402.00782.

\bibitem[{{Chen} et~al.(2024){Chen}, {Zhou}, {Zhao}, {Wang}, and {Wen}}]{chen2024MathField}
Zhipeng {Chen}, Kun {Zhou}, Wayne~Xin {Zhao}, Jingyuan {Wang}, and Ji-Rong {Wen}. 2024.
\newblock \href {https://doi.org/10.48550/arXiv.2406.12606} {{Not Everything is All You Need: Toward Low-Redundant Optimization for Large Language Model Alignment}}.
\newblock \emph{arXiv e-prints}, arXiv:2406.12606.

\bibitem[{{Cheng} et~al.(2025){Cheng}, {Huang}, {Zhu}, {Dai}, {Zhao}, {Zhang}, and {Wei}}]{cheng2025Reasoning_with_Exploration}
Daixuan {Cheng}, Shaohan {Huang}, Xuekai {Zhu}, Bo~{Dai}, Wayne~Xin {Zhao}, Zhenliang {Zhang}, and Furu {Wei}. 2025.
\newblock \href {https://doi.org/10.48550/arXiv.2506.14758} {{Reasoning with Exploration: An Entropy Perspective on Reinforcement Learning for LLMs}}.
\newblock \emph{arXiv e-prints}, arXiv:2506.14758.

\bibitem[{{Cui} et~al.(2025){Cui}, {Zhang}, {Chen}, {Yuan}, {Wang}, {Zuo}, {Li}, {Fan}, {Chen}, {Chen}, {Liu}, {Peng}, {Bai}, {Ouyang}, {Cheng}, {Zhou}, and {Ding}}]{cui2025entropyMechanism}
Ganqu {Cui}, Yuchen {Zhang}, Jiacheng {Chen}, Lifan {Yuan}, Zhi {Wang}, Yuxin {Zuo}, Haozhan {Li}, Yuchen {Fan}, Huayu {Chen}, Weize {Chen}, Zhiyuan {Liu}, Hao {Peng}, Lei {Bai}, Wanli {Ouyang}, Yu~{Cheng}, Bowen {Zhou}, and Ning {Ding}. 2025.
\newblock \href {https://doi.org/10.48550/arXiv.2505.22617} {{The Entropy Mechanism of Reinforcement Learning for Reasoning Language Models}}.
\newblock \emph{arXiv e-prints}, arXiv:2505.22617.

\bibitem[{{Da} et~al.(2025){Da}, {Wang}, {Deng}, {Ma}, {Barhate}, and {Hendryx}}]{da2025program}
Jeff {Da}, Clinton {Wang}, Xiang {Deng}, Yuntao {Ma}, Nikhil {Barhate}, and Sean {Hendryx}. 2025.
\newblock \href {https://doi.org/10.48550/arXiv.2506.11425} {{Agent-RLVR: Training Software Engineering Agents via Guidance and Environment Rewards}}.
\newblock \emph{arXiv e-prints}, arXiv:2506.11425.

\bibitem[{{DeepSeek-AI} et~al.(2025){DeepSeek-AI}, {Guo}, {Yang}, {Zhang}, {Song}, {Zhang}, {Xu}, {Zhu}, {Ma}, {Wang}, {Bi}, {Zhang}, {Yu}, {Wu}, {Wu}, {Gou}, {Shao}, {Li}, {Gao}, {Liu}, {Xue}, {Wang}, {Wu}, {Feng}, {Lu}, {Zhao}, {Deng}, {Zhang}, {Ruan}, {Dai}, {Chen}, {Ji}, {Li}, {Lin}, {Dai}, {Luo}, {Hao}, {Chen}, {Li}, {Zhang}, {Bao}, {Xu}, {Wang}, {Ding}, {Xin}, {Gao}, {Qu}, {Li}, {Guo}, {Li}, {Wang}, {Chen}, {Yuan}, {Qiu}, {Li}, {Cai}, {Ni}, {Liang}, {Chen}, {Dong}, {Hu}, {Gao}, {Guan}, {Huang}, {Yu}, {Wang}, {Zhang}, {Zhao}, {Wang}, {Zhang}, {Xu}, {Xia}, {Zhang}, {Zhang}, {Tang}, {Li}, {Wang}, {Li}, {Tian}, {Huang}, {Zhang}, {Wang}, {Chen}, {Du}, {Ge}, {Zhang}, {Pan}, {Wang}, {Chen}, {Jin}, {Chen}, {Lu}, {Zhou}, {Chen}, {Ye}, {Wang}, {Yu}, {Zhou}, {Pan}, {Li}, {Zhou}, {Wu}, {Ye}, {Yun}, {Pei}, {Sun}, {Wang}, {Zeng}, {Zhao}, {Liu}, {Liang}, {Gao}, {Yu}, {Zhang}, {Xiao}, {An}, {Liu}, {Wang}, {Chen}, {Nie}, {Cheng}, {Liu}, {Xie}, {Liu}, {Yang}, {Li}, {Su}, {Lin}, {Li}, {Jin}, {Shen}, {Chen}, {Sun}, {Wang},
  {Song}, {Zhou}, {Wang}, {Shan}, {Li}, {Wang}, {Wei}, {Zhang}, {Xu}, {Li}, {Zhao}, {Sun}, {Wang}, {Yu}, {Zhang}, {Shi}, {Xiong}, {He}, {Piao}, {Wang}, {Tan}, {Ma}, {Liu}, {Guo}, {Ou}, {Wang}, {Gong}, {Zou}, {He}, {Xiong}, {Luo}, {You}, {Liu}, {Zhou}, {Zhu}, {Xu}, {Huang}, {Li}, {Zheng}, {Zhu}, {Ma}, {Tang}, {Zha}, {Yan}, {Ren}, {Ren}, {Sha}, {Fu}, {Xu}, {Xie}, {Zhang}, {Hao}, {Ma}, {Yan}, {Wu}, {Gu}, {Zhu}, {Liu}, {Li}, {Xie}, {Song}, {Pan}, {Huang}, {Xu}, {Zhang}, and {Zhang}}]{guo2025distill}
{DeepSeek-AI}, Daya {Guo}, Dejian {Yang}, Haowei {Zhang}, Junxiao {Song}, Ruoyu {Zhang}, Runxin {Xu}, Qihao {Zhu}, Shirong {Ma}, Peiyi {Wang}, Xiao {Bi}, Xiaokang {Zhang}, Xingkai {Yu}, Yu~{Wu}, Z.~F. {Wu}, Zhibin {Gou}, Zhihong {Shao}, Zhuoshu {Li}, Ziyi {Gao}, and 181 others. 2025.
\newblock \href {https://doi.org/10.48550/arXiv.2501.12948} {{DeepSeek-R1: Incentivizing Reasoning Capability in LLMs via Reinforcement Learning}}.
\newblock \emph{arXiv e-prints}, arXiv:2501.12948.

\bibitem[{{Guo} et~al.(2025){Guo}, {Xu}, {Liu}, {Ye}, and {Qiu}}]{guo2025SPO}
Yiran {Guo}, Lijie {Xu}, Jie {Liu}, Dan {Ye}, and Shuang {Qiu}. 2025.
\newblock \href {https://doi.org/10.48550/arXiv.2505.23564} {{Segment Policy Optimization: Effective Segment-Level Credit Assignment in RL for Large Language Models}}.
\newblock \emph{arXiv e-prints}, arXiv:2505.23564.

\bibitem[{{He} et~al.(2024){He}, {Luo}, {Bai}, {Hu}, {Leng Thai}, {Shen}, {Hu}, {Han}, {Huang}, {Zhang}, {Liu}, {Qi}, {Liu}, and {Sun}}]{He2024Olympiad}
Chaoqun {He}, Renjie {Luo}, Yuzhuo {Bai}, Shengding {Hu}, Zhen {Leng Thai}, Junhao {Shen}, Jinyi {Hu}, Xu~{Han}, Yujie {Huang}, Yuxiang {Zhang}, Jie {Liu}, Lei {Qi}, Zhiyuan {Liu}, and Maosong {Sun}. 2024.
\newblock \href {https://doi.org/10.48550/arXiv.2402.14008} {{OlympiadBench: A Challenging Benchmark for Promoting AGI with Olympiad-Level Bilingual Multimodal Scientific Problems}}.
\newblock \emph{arXiv e-prints}, arXiv:2402.14008.

\bibitem[{{Hendrycks} et~al.(2021){Hendrycks}, {Burns}, {Kadavath}, {Arora}, {Basart}, {Tang}, {Song}, and {Steinhardt}}]{hendrycks2021Math}
Dan {Hendrycks}, Collin {Burns}, Saurav {Kadavath}, Akul {Arora}, Steven {Basart}, Eric {Tang}, Dawn {Song}, and Jacob {Steinhardt}. 2021.
\newblock \href {https://doi.org/10.48550/arXiv.2103.03874} {{Measuring Mathematical Problem Solving With the MATH Dataset}}.
\newblock \emph{arXiv e-prints}, arXiv:2103.03874.

\bibitem[{{Jing} et~al.(2025){Jing}, {Lee}, {Zhang}, {Zhou}, {Yuan}, {Gao}, {Zhu}, {Papanastasiou}, {Fang}, and {Yang}}]{jing2025science}
Peiyuan {Jing}, Kinhei {Lee}, Zhenxuan {Zhang}, Huichi {Zhou}, Zhengqing {Yuan}, Zhifan {Gao}, Lei {Zhu}, Giorgos {Papanastasiou}, Yingying {Fang}, and Guang {Yang}. 2025.
\newblock \href {https://doi.org/10.48550/arXiv.2504.18453} {{Reason Like a Radiologist: Chain-of-Thought and Reinforcement Learning for Verifiable Report Generation}}.
\newblock \emph{arXiv e-prints}, arXiv:2504.18453.

\bibitem[{{Lewkowycz} et~al.(2022){Lewkowycz}, {Andreassen}, {Dohan}, {Dyer}, {Michalewski}, {Ramasesh}, {Slone}, {Anil}, {Schlag}, {Gutman-Solo}, {Wu}, {Neyshabur}, {Gur-Ari}, and {Misra}}]{lew2022Minerva}
Aitor {Lewkowycz}, Anders {Andreassen}, David {Dohan}, Ethan {Dyer}, Henryk {Michalewski}, Vinay {Ramasesh}, Ambrose {Slone}, Cem {Anil}, Imanol {Schlag}, Theo {Gutman-Solo}, Yuhuai {Wu}, Behnam {Neyshabur}, Guy {Gur-Ari}, and Vedant {Misra}. 2022.
\newblock \href {https://doi.org/10.48550/arXiv.2206.14858} {{Solving Quantitative Reasoning Problems with Language Models}}.
\newblock \emph{arXiv e-prints}, arXiv:2206.14858.

\bibitem[{LI et~al.(2024)LI, Beeching, Tunstall, Lipkin, Soletskyi, Huang, Rasul, Yu, Jiang, Shen, Qin, Dong, Zhou, Fleureau, Lample, and Polu}]{li2024AIME}
Jia LI, Edward Beeching, Lewis Tunstall, Ben Lipkin, Roman Soletskyi, Shengyi~Costa Huang, Kashif Rasul, Longhui Yu, Albert Jiang, Ziju Shen, Zihan Qin, Bin Dong, Li~Zhou, Yann Fleureau, Guillaume Lample, and Stanislas Polu. 2024.
\newblock Numinamath.
\newblock \url{[https://huggingface.co/AI-MO/NuminaMath-CoT](https://github.com/project-numina/aimo-progress-prize/blob/main/report/numina_dataset.pdf)}.

\bibitem[{{Lin} et~al.(2024){Lin}, {Liang}, {Xu}, {Lin}, {Wang}, {Luo}, {Shi}, {Li}, {Yang}, and {Tu}}]{lincritical}
Zicheng {Lin}, Tian {Liang}, Jiahao {Xu}, Qiuzhi {Lin}, Xing {Wang}, Ruilin {Luo}, Chufan {Shi}, Siheng {Li}, Yujiu {Yang}, and Zhaopeng {Tu}. 2024.
\newblock \href {https://doi.org/10.48550/arXiv.2411.19943} {{Critical Tokens Matter: Token-Level Contrastive Estimation Enhances LLM's Reasoning Capability}}.
\newblock \emph{arXiv e-prints}, arXiv:2411.19943.

\bibitem[{{Liu} et~al.(2025){Liu}, {Gong}, {Fu}, {Liu}, {Chen}, {Hu}, {Zhang}, {Liu}, {Zhang}, and {Tu}}]{liu2025GHPO}
Ziru {Liu}, Cheng {Gong}, Xinyu {Fu}, Yaofang {Liu}, Ran {Chen}, Shoubo {Hu}, Suiyun {Zhang}, Rui {Liu}, Qingfu {Zhang}, and Dandan {Tu}. 2025.
\newblock \href {https://doi.org/10.48550/arXiv.2507.10628} {{GHPO: Adaptive Guidance for Stable and Efficient LLM Reinforcement Learning}}.
\newblock \emph{arXiv e-prints}, arXiv:2507.10628.

\bibitem[{{M2 Team} et~al.(2025){M2 Team}, {Dou}, {Liu}, {Yang}, {Li}, {Jia}, {Chen}, {Ju}, {Wang}, {Dang}, {Li}, {Zeng}, {Zhou}, {Zhu}, {Pan}, {Deng}, {Ai}, {Dong}, {Zhang}, {Tai}, {Hong}, {Lu}, {Sun}, {Guo}, {Ma}, {Xin}, {Yang}, {Zhang}, {Mo}, {Liang}, {Zhang}, {Cui}, {Zhu}, and {Wang}}]{baichuan2025M2}
{M2 Team}, Chengfeng {Dou}, Chong {Liu}, Fan {Yang}, Fei {Li}, Jiyuan {Jia}, Mingyang {Chen}, Qiang {Ju}, Shuai {Wang}, Shunya {Dang}, Tianpeng {Li}, Xiangrong {Zeng}, Yijie {Zhou}, Chenzheng {Zhu}, Da~{Pan}, Fei {Deng}, Guangwei {Ai}, Guosheng {Dong}, Hongda {Zhang}, and 15 others. 2025.
\newblock \href {https://doi.org/10.48550/arXiv.2509.02208} {{Baichuan-M2: Scaling Medical Capability with Large Verifier System}}.
\newblock \emph{arXiv e-prints}, arXiv:2509.02208.

\bibitem[{{Qwen} et~al.(2024){Qwen}, {:}, {Yang}, {Yang}, {Zhang}, {Hui}, {Zheng}, {Yu}, {Li}, {Liu}, {Huang}, {Wei}, {Lin}, {Yang}, {Tu}, {Zhang}, {Yang}, {Yang}, {Zhou}, {Lin}, {Dang}, {Lu}, {Bao}, {Yang}, {Yu}, {Li}, {Xue}, {Zhang}, {Zhu}, {Men}, {Lin}, {Li}, {Tang}, {Xia}, {Ren}, {Ren}, {Fan}, {Su}, {Zhang}, {Wan}, {Liu}, {Cui}, {Zhang}, and {Qiu}}]{qwen2024qwen}
{Qwen}, {:}, An~{Yang}, Baosong {Yang}, Beichen {Zhang}, Binyuan {Hui}, Bo~{Zheng}, Bowen {Yu}, Chengyuan {Li}, Dayiheng {Liu}, Fei {Huang}, Haoran {Wei}, Huan {Lin}, Jian {Yang}, Jianhong {Tu}, Jianwei {Zhang}, Jianxin {Yang}, Jiaxi {Yang}, Jingren {Zhou}, and 25 others. 2024.
\newblock \href {https://doi.org/10.48550/arXiv.2412.15115} {{Qwen2.5 Technical Report}}.
\newblock \emph{arXiv e-prints}, arXiv:2412.15115.

\bibitem[{{Sellergren} et~al.(2025){Sellergren}, {Kazemzadeh}, {Jaroensri}, {Kiraly}, {Traverse}, {Kohlberger}, {Xu}, {Jamil}, {Hughes}, {Lau}, {Chen}, {Mahvar}, {Yatziv}, {Chen}, {Sterling}, {Baby}, {Baby}, {Lai}, {Schmidgall}, {Yang}, {Chen}, {Bjornsson}, {Reddy}, {Brush}, {Philbrick}, {Asiedu}, {Mezerreg}, {Hu}, {Yang}, {Tiwari}, {Jansen}, {Singh}, {Liu}, {Azizi}, {Kamath}, {Ferret}, {Pathak}, {Vieillard}, {Merhej}, {Perrin}, {Matejovicova}, {Ram{\'e}}, {Riviere}, {Rouillard}, {Mesnard}, {Cideron}, {Grill}, {Ramos}, {Yvinec}, {Casbon}, {Buchatskaya}, {Alayrac}, {Lepikhin}, {Feinberg}, {Borgeaud}, {Andreev}, {Hardin}, {Dadashi}, {Hussenot}, {Joulin}, {Bachem}, {Matias}, {Chou}, {Hassidim}, {Goel}, {Farabet}, {Barral}, {Warkentin}, {Shlens}, {Fleet}, {Cotruta}, {Sanseviero}, {Martins}, {Kirk}, {Rao}, {Shetty}, {Steiner}, {Kirmizibayrak}, {Pilgrim}, {Golden}, and {Yang}}]{Sellergren2025science}
Andrew {Sellergren}, Sahar {Kazemzadeh}, Tiam {Jaroensri}, Atilla {Kiraly}, Madeleine {Traverse}, Timo {Kohlberger}, Shawn {Xu}, Fayaz {Jamil}, C{\'\i}an {Hughes}, Charles {Lau}, Justin {Chen}, Fereshteh {Mahvar}, Liron {Yatziv}, Tiffany {Chen}, Bram {Sterling}, Stefanie~Anna {Baby}, Susanna~Maria {Baby}, Jeremy {Lai}, Samuel {Schmidgall}, and 62 others. 2025.
\newblock \href {https://doi.org/10.48550/arXiv.2507.05201} {{MedGemma Technical Report}}.
\newblock \emph{arXiv e-prints}, arXiv:2507.05201.

\bibitem[{{Shao} et~al.(2024){Shao}, {Wang}, {Zhu}, {Xu}, {Song}, {Bi}, {Zhang}, {Zhang}, {Li}, {Wu}, and {Guo}}]{shao2024grpo}
Zhihong {Shao}, Peiyi {Wang}, Qihao {Zhu}, Runxin {Xu}, Junxiao {Song}, Xiao {Bi}, Haowei {Zhang}, Mingchuan {Zhang}, Y.~K. {Li}, Y.~{Wu}, and Daya {Guo}. 2024.
\newblock \href {https://doi.org/10.48550/arXiv.2402.03300} {{DeepSeekMath: Pushing the Limits of Mathematical Reasoning in Open Language Models}}.
\newblock \emph{arXiv e-prints}, arXiv:2402.03300.

\bibitem[{{Shen}(2025)}]{Shen2025Aent}
Han {Shen}. 2025.
\newblock \href {https://doi.org/10.48550/arXiv.2509.03493} {{On Entropy Control in LLM-RL Algorithms}}.
\newblock \emph{arXiv e-prints}, arXiv:2509.03493.

\bibitem[{{Sheng} et~al.(2024){Sheng}, {Zhang}, {Ye}, {Wu}, {Zhang}, {Zhang}, {Peng}, {Lin}, and {Wu}}]{sheng2024Verl}
Guangming {Sheng}, Chi {Zhang}, Zilingfeng {Ye}, Xibin {Wu}, Wang {Zhang}, Ru~{Zhang}, Yanghua {Peng}, Haibin {Lin}, and Chuan {Wu}. 2024.
\newblock \href {https://doi.org/10.48550/arXiv.2409.19256} {{HybridFlow: A Flexible and Efficient RLHF Framework}}.
\newblock \emph{arXiv e-prints}, arXiv:2409.19256.

\bibitem[{Vassoyan et~al.(2025)Vassoyan, Beau, and Plaud}]{vassoyan}
Jean Vassoyan, Nathana{\"e}l Beau, and Roman Plaud. 2025.
\newblock \href {https://doi.org/10.18653/v1/2025.findings-naacl.340} {Ignore the {KL} penalty! boosting exploration on critical tokens to enhance {RL} fine-tuning}.
\newblock In \emph{Findings of the Association for Computational Linguistics: NAACL 2025}, pages 6108--6118, Albuquerque, New Mexico. Association for Computational Linguistics.

\bibitem[{{Wang} et~al.(2025){Wang}, {Yu}, {Gao}, {Zheng}, {Liu}, {Lu}, {Dang}, {Chen}, {Yang}, {Zhang}, {Liu}, {Yang}, {Zhao}, {Yue}, {Song}, {Yu}, {Huang}, and {Lin}}]{wang2025_8020}
Shenzhi {Wang}, Le~{Yu}, Chang {Gao}, Chujie {Zheng}, Shixuan {Liu}, Rui {Lu}, Kai {Dang}, Xionghui {Chen}, Jianxin {Yang}, Zhenru {Zhang}, Yuqiong {Liu}, An~{Yang}, Andrew {Zhao}, Yang {Yue}, Shiji {Song}, Bowen {Yu}, Gao {Huang}, and Junyang {Lin}. 2025.
\newblock \href {https://doi.org/10.48550/arXiv.2506.01939} {{Beyond the 80/20 Rule: High-Entropy Minority Tokens Drive Effective Reinforcement Learning for LLM Reasoning}}.
\newblock \emph{arXiv e-prints}, arXiv:2506.01939.

\bibitem[{{Wei} et~al.(2025{\natexlab{a}}){Wei}, {Zeng}, {Li}, {Brown}, {Frunza}, {Deng}, {Schneider}, {Nevmyvaka}, {Zhao}, {Garcia}, and {Hong}}]{wei2025CreaditAssign}
Quan {Wei}, Siliang {Zeng}, Chenliang {Li}, William {Brown}, Oana {Frunza}, Wei {Deng}, Anderson {Schneider}, Yuriy {Nevmyvaka}, Yang~Katie {Zhao}, Alfredo {Garcia}, and Mingyi {Hong}. 2025{\natexlab{a}}.
\newblock \href {https://doi.org/10.48550/arXiv.2505.11821} {{Reinforcing Multi-Turn Reasoning in LLM Agents via Turn-Level Reward Design}}.
\newblock \emph{arXiv e-prints}, arXiv:2505.11821.

\bibitem[{{Wei} et~al.(2025{\natexlab{b}}){Wei}, {Duchenne}, {Copet}, {Carbonneaux}, {Zhang}, {Fried}, {Synnaeve}, {Singh}, and {Wang}}]{wei2025SWE-RL}
Yuxiang {Wei}, Olivier {Duchenne}, Jade {Copet}, Quentin {Carbonneaux}, Lingming {Zhang}, Daniel {Fried}, Gabriel {Synnaeve}, Rishabh {Singh}, and Sida~I. {Wang}. 2025{\natexlab{b}}.
\newblock \href {https://doi.org/10.48550/arXiv.2502.18449} {{SWE-RL: Advancing LLM Reasoning via Reinforcement Learning on Open Software Evolution}}.
\newblock \emph{arXiv e-prints}, arXiv:2502.18449.

\bibitem[{{Xie} et~al.(2025){Xie}, {Shi}, {Tian}, {Yao}, and {Zhang}}]{xie2025capo}
Guofu {Xie}, Yunsheng {Shi}, Hongtao {Tian}, Ting {Yao}, and Xiao {Zhang}. 2025.
\newblock \href {https://doi.org/10.48550/arXiv.2508.02298} {{CAPO: Towards Enhancing LLM Reasoning through Generative Credit Assignment}}.
\newblock \emph{arXiv e-prints}, arXiv:2508.02298.

\bibitem[{{Yang} et~al.(2024){Yang}, {Zhang}, {Hui}, {Gao}, {Yu}, {Li}, {Liu}, {Tu}, {Zhou}, {Lin}, {Lu}, {Xue}, {Lin}, {Liu}, {Ren}, and {Zhang}}]{yang2024Qwen2.5-Math}
An~{Yang}, Beichen {Zhang}, Binyuan {Hui}, Bofei {Gao}, Bowen {Yu}, Chengpeng {Li}, Dayiheng {Liu}, Jianhong {Tu}, Jingren {Zhou}, Junyang {Lin}, Keming {Lu}, Mingfeng {Xue}, Runji {Lin}, Tianyu {Liu}, Xingzhang {Ren}, and Zhenru {Zhang}. 2024.
\newblock \href {https://doi.org/10.48550/arXiv.2409.12122} {{Qwen2.5-Math Technical Report: Toward Mathematical Expert Model via Self-Improvement}}.
\newblock \emph{arXiv e-prints}, arXiv:2409.12122.

\bibitem[{{Yang} et~al.(2025){Yang}, {chen}, {Wang}, and {Li}}]{yang2025SSPO}
Kun {Yang}, Zikang {chen}, Yanmeng {Wang}, and Zhigen {Li}. 2025.
\newblock \href {https://arxiv.org/abs/2511.04256} {{SSPO: Subsentence-level Policy Optimization}}.
\newblock \emph{arXiv e-prints}, arXiv:2511.04256.

\bibitem[{{Yu} et~al.(2025){Yu}, {Zhang}, {Zhu}, {Yuan}, {Zuo}, {Yue}, {Dai}, {Fan}, {Liu}, {Liu}, {Liu}, {Lin}, {Lin}, {Ma}, {Sheng}, {Tong}, {Zhang}, {Zhang}, {Zhang}, {Zhu}, {Zhu}, {Chen}, {Chen}, {Wang}, {Yu}, {Song}, {Wei}, {Zhou}, {Liu}, {Ma}, {Zhang}, {Yan}, {Qiao}, {Wu}, and {Wang}}]{yu2025dapo}
Qiying {Yu}, Zheng {Zhang}, Ruofei {Zhu}, Yufeng {Yuan}, Xiaochen {Zuo}, Yu~{Yue}, Weinan {Dai}, Tiantian {Fan}, Gaohong {Liu}, Lingjun {Liu}, Xin {Liu}, Haibin {Lin}, Zhiqi {Lin}, Bole {Ma}, Guangming {Sheng}, Yuxuan {Tong}, Chi {Zhang}, Mofan {Zhang}, Wang {Zhang}, and 16 others. 2025.
\newblock \href {https://doi.org/10.48550/arXiv.2503.14476} {{DAPO: An Open-Source LLM Reinforcement Learning System at Scale}}.
\newblock \emph{arXiv e-prints}, arXiv:2503.14476.

\bibitem[{{Yue} et~al.(2025){Yue}, {Yuan}, {Yu}, {Zuo}, {Zhu}, {Xu}, {Chen}, {Wang}, {Fan}, {Du}, {Wei}, {Yu}, {Liu}, {Liu}, {Liu}, {Lin}, {Lin}, {Ma}, {Zhang}, {Zhang}, {Zhang}, {Zhu}, {Zhang}, {Liu}, {Wang}, {Wu}, and {Yan}}]{yue2025vapo}
Yu~{Yue}, Yufeng {Yuan}, Qiying {Yu}, Xiaochen {Zuo}, Ruofei {Zhu}, Wenyuan {Xu}, Jiaze {Chen}, Chengyi {Wang}, TianTian {Fan}, Zhengyin {Du}, Xiangpeng {Wei}, Xiangyu {Yu}, Gaohong {Liu}, Juncai {Liu}, Lingjun {Liu}, Haibin {Lin}, Zhiqi {Lin}, Bole {Ma}, Chi {Zhang}, and 8 others. 2025.
\newblock \href {https://doi.org/10.48550/arXiv.2504.05118} {{VAPO: Efficient and Reliable Reinforcement Learning for Advanced Reasoning Tasks}}.
\newblock \emph{arXiv e-prints}, arXiv:2504.05118.

\bibitem[{Zhang et~al.(2025{\natexlab{a}})Zhang, Xie, Yu, Xu, Tang, Li, and Xu}]{legal2}
Kepu Zhang, Guofu Xie, Weijie Yu, Mingyue Xu, Xu~Tang, Yaxin Li, and Jun Xu. 2025{\natexlab{a}}.
\newblock \href {https://doi.org/10.18653/v1/2025.findings-emnlp.84} {Legal mathematical reasoning with {LLM}s: Procedural alignment through two-stage reinforcement learning}.
\newblock In \emph{Findings of the Association for Computational Linguistics: EMNLP 2025}, pages 1586--1598, Suzhou, China. Association for Computational Linguistics.

\bibitem[{Zhang et~al.(2025{\natexlab{b}})Zhang, Yu, Sun, and Xu}]{legal1}
Kepu Zhang, Weijie Yu, Zhongxiang Sun, and Jun Xu. 2025{\natexlab{b}}.
\newblock \href {https://doi.org/10.1145/3746252.3761120} {Syler: A framework for explicit syllogistic legal reasoning in large language models}.
\newblock In \emph{Proceedings of the 34th ACM International Conference on Information and Knowledge Management}, CIKM '25, page 4117–4127, New York, NY, USA. Association for Computing Machinery.

\bibitem[{{Zhang} et~al.(2025){Zhang}, {Wen}, {Wu}, and {Huang}}]{zhang2025EDGE-GRPO}
Xingjian {Zhang}, Siwei {Wen}, Wenjun {Wu}, and Lei {Huang}. 2025.
\newblock \href {https://doi.org/10.48550/arXiv.2507.21848} {{EDGE-GRPO: Entropy-Driven GRPO with Guided Error Correction for Advantage Diversity}}.
\newblock \emph{arXiv e-prints}, arXiv:2507.21848.

\bibitem[{{Zheng} et~al.(2025){Zheng}, {Xing}, {Gu}, {Liang}, {Qu}, {Zhou}, {Li}, {Wen}, {Lin}, {Huang}, {Liu}, {Zhang}, and {Ma}}]{zheng2025FR3E}
Tianyu {Zheng}, Tianshun {Xing}, Qingshui {Gu}, Taoran {Liang}, Xingwei {Qu}, Xin {Zhou}, Yizhi {Li}, Zhoufutu {Wen}, Chenghua {Lin}, Wenhao {Huang}, Qian {Liu}, Ge~{Zhang}, and Zejun {Ma}. 2025.
\newblock \href {https://doi.org/10.48550/arXiv.2507.07017} {{First Return, Entropy-Eliciting Explore}}.
\newblock \emph{arXiv e-prints}, arXiv:2507.07017.

\end{thebibliography}
\clearpage
\appendix

\section{Appendix}
\label{sec:appendix}
\subsection{Algorithm and Time Complexity Analysis}
\label{app:less-algorithm}
\begin{algorithm}[H]
\caption{LESS: Low-Entropy Segment Shaping}
\label{LESS_algorithm}
\begin{algorithmic}[1]
\STATE \textbf{Input:}
\STATE ~~Group of responses $\mathcal{G}=\{O_1,\dots,O_G\}$,
token advantages $\mathcal{A}=\{A_j^i \mid t_j\in O_i,\, i=1,\dots,G\}$,
token entropies $H=\{\mathcal{H}_j^i \mid t_j\in O_i,\, i=1,\dots,G\}$,
correctness labels $\{correct_1,\dots,correct_G\}$,
entropy quantile $h$,
minimum segment length $\mu$.
\STATE \textbf{Output:} Shaped advantages $\mathcal{A}'=\{\hat{A}_j^i \mid t_j\in O_i,\, i=1,\dots,G\}$

\STATE $N_r \gets \sum_i \mathbb{I}[correct_i = 1]$;\quad
       $N_w \gets \sum_i \mathbb{I}[correct_i = 0]$
\FOR{each response $O_i$}
    \STATE Compute entropy threshold $\tau_i$ from $\{\mathcal{H}_j^i\}_{t_j\in O_i}$ using quantile $h$
    \STATE Segment $O_i$ into $\mathcal{S}_i^{\text{high}}$, $\mathcal{S}_i^{\text{frag}}$, $\mathcal{S}_i^{\text{seg}}$ using $\tau_i$ and $\mu$ (Eq.~\eqref{eq:segment-sets})
\ENDFOR

\STATE $\Sigma \gets \varnothing$ \COMMENT{set of unique low-entropy segments}
\FOR{each response $O_i$}
    \FORALL{$\sigma\in \mathcal{S}_i^{\text{seg}}$}
        \IF{no $\sigma'\in\Sigma$ is a contiguous segments of $\sigma$}
            \STATE $\Sigma \gets \Sigma \cup \{\sigma\}$
            \STATE Remove from $\Sigma$ any $\sigma'$ that is strictly contained in $\sigma$
        \ENDIF
    \ENDFOR
\ENDFOR

\FORALL{$\sigma\in\Sigma$}
    \STATE Compute $n_r(\sigma)$ and $n_w(\sigma)$ according to Eq.~\eqref{eq:count}
\ENDFOR

\FOR{each response $O_i$}
    \FOR{each token $t_j \in O_i$}
        \STATE Set $\hat{A}_j^i$ according to Eq.~\eqref{eq:shaping}
    \ENDFOR
\ENDFOR

\RETURN $\mathcal{A}'=\{\hat{A}_j^i\}$
\end{algorithmic}
\end{algorithm}
The overall LESS procedure is summarized in Algorithm~\ref{LESS_algorithm}.
Given a group of responses and their token-level entropies, we first compute
an entropy threshold for each response and segment it into high-entropy tokens,
short low-entropy fragments, and longer low-entropy segments
(Eq.~\eqref{eq:segment-sets}). We then build a set $\Sigma$ of non-redundant
low-entropy segments across the group by keeping only segments that are not
strictly contained in longer ones. For each segment $\sigma\in\Sigma$, we
count how many correct and incorrect responses it appears in
(Eq.~\eqref{eq:count}), and finally assign a shaped advantage to every token
based on its entropy category and the statistics of the segment it belongs to
(Eq.~\eqref{eq:shaping}). The resulting token-level advantages
$\mathcal{A}'$ can be plugged into any group-based RLVR update.

In terms of complexity, when the batch size is $B$, the group size is $G$,
and the maximum response length is $L$, the segmentation and shaping
operations visit each token a constant number of times, giving a practical
time complexity of $O(BGL)$. Under our main setting (batch size $512$,
group size $8$, average response length about $800$), LESS adds roughly
60 seconds of overhead in our implementation, which is small compared to
the overall RL training time.

\subsection{Prompt Template}
\label{app:prompt-template}
We use a unified prompt template for all training and evaluation experiments, adapted from the official Qwen-Math template~\citep{yang2024Qwen2.5-Math}. The concrete format is shown in Table~\ref{tab:prompt-template}.

\begin{table}[t]
\centering
\small
\caption{Prompt template used for all experiments. \texttt{\{question\}} is replaced by the problem description.}
\begin{tabular}{p{0.9\linewidth}}
\toprule
\textbf{Prompt Template} \\
\midrule
\texttt{<|im start|>system} \\
\texttt{Please reason step by step, and put your final answer within \textbackslash boxed\{\}.} \\
\texttt{<|im end|>} \\
\texttt{<|im start|>user} \\
\texttt{\{question\}} \\
\texttt{<|im end|>} \\
\texttt{<|im start|>assistant} \\
\bottomrule
\end{tabular}

\label{tab:prompt-template}
\end{table}

\subsection{Additional Overlap–Accuracy Curves}
\label{app:curves}
\begin{figure}[ht]
  \begin{subfigure}{\linewidth}
    \centering
    \includegraphics[width=1\linewidth]{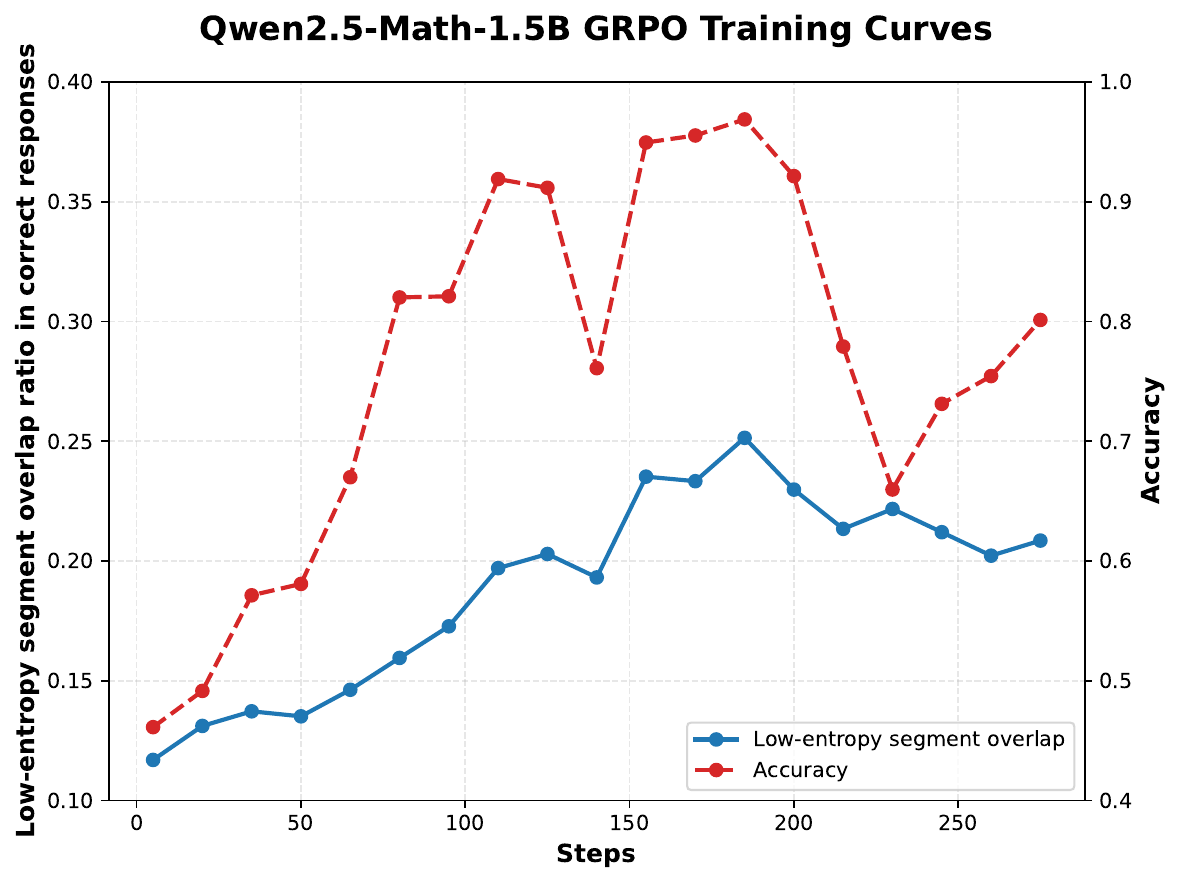}
    \subcaption{Qwen2.5-1.5B-Math}
    \label{sub:qwen7b-math}
  \end{subfigure}
  \begin{subfigure}{\linewidth}
    \centering
    \includegraphics[width=1\linewidth]{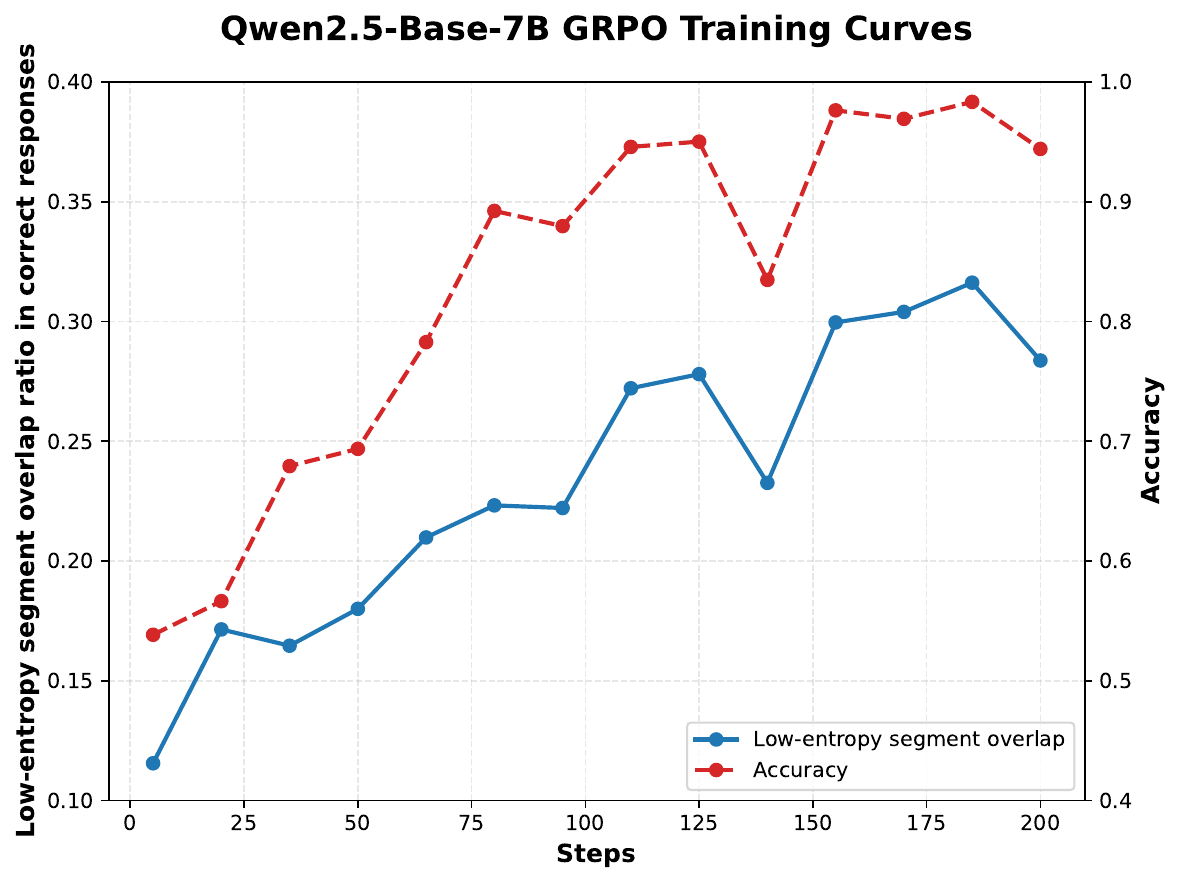}
    \subcaption{Qwen2.5-Base-7B}
    \label{sub:qwen7b-base}
  \end{subfigure}
  \caption {GRPO training curves on two additional backbones. 
We plot the overlap ratio of low-entropy segments in correct responses (left y-axis, blue) and accuracy (right y-axis, red) over training steps for 
(a) Qwen2.5-Math-1.5B and 
(b) Qwen2.5-Base-7B. 
In both models, low-entropy overlap and accuracy increase in tandem, echoing the trend observed for Qwen2.5-Math-7B in the main text (\ref{fig:fig1}).}
  \label{fig:trend_2}
\end{figure}
To check whether the correlation between low-entropy overlap and accuracy holds beyond Qwen2.5-Math-7B, 
we repeat the analysis in §~\ref{sec:preliminary} on Qwen2.5-Math-1.5B and Qwen2.5-Base-7B. 
Figure~\ref{fig:trend_2} shows that, under GRPO training, the overlap ratio of low-entropy segments in correct responses grows together with accuracy for both backbones. 
Early in training, both curves are low and noisy; as learning proceeds, the overlap becomes higher and smoother while accuracy also rises. These results support our claim that low-entropy segment overlap tracks the formation of stable reasoning routines across different model sizes and pretraining setups.
\subsection{Additional Accuracy–Overlap Correlations}
\label{app:extra-corr}

In §~\ref{sec:preliminary}, we report that, for Qwen2.5-Math-7B, benchmark accuracy
is strongly correlated with the overlap of low-entropy segments across
correct responses. To test the robustness of this phenomenon, we repeat
the correlation analysis on four additional backbones:
Qwen2.5-7B-Base, DeepSeek-R1-Distill-Llama-8B,
DeepSeek-R1-Distill-Qwen-7B, and Qwen2.5-Math-1.5B-Oat-Zero.

\begin{figure*}[t]
  \begin{subfigure}{1\linewidth}
    \includegraphics[width=1\linewidth]{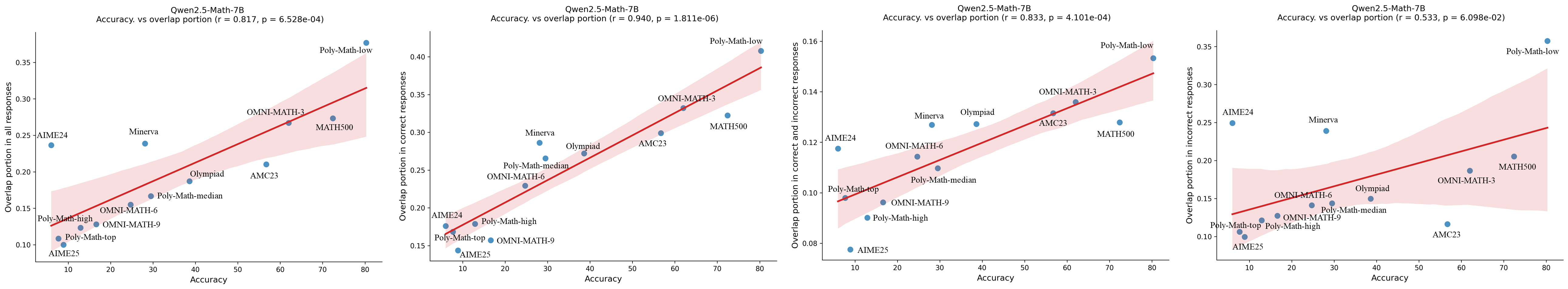}
    \subcaption{Qwen2.5-Math-7B}
    \label{sub:qwen7b-math}
  \end{subfigure}
  \\[5pt] 
  
  \begin{subfigure}{1\linewidth}
    \includegraphics[width=1\linewidth]{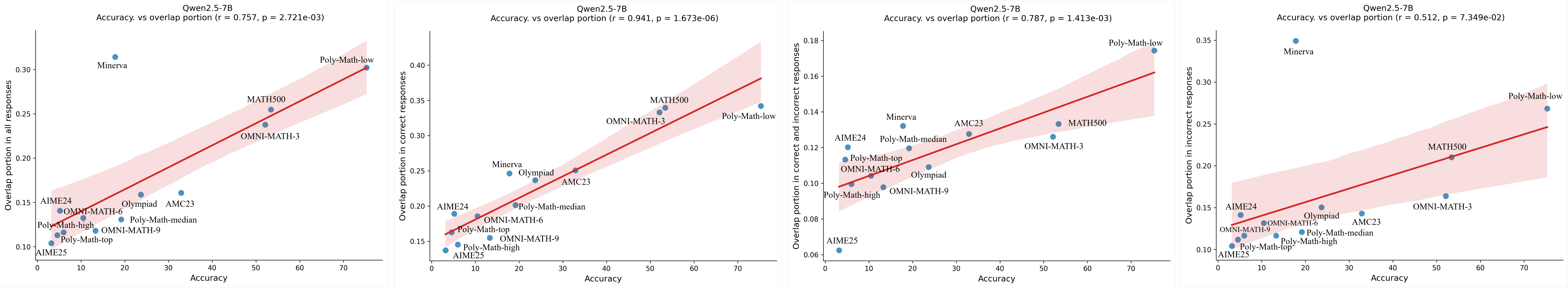}
    \subcaption{Qwen2.5-7B-Base}
    \label{sub:qwen7b-base}
  \end{subfigure}
  \\[5pt]
  \begin{subfigure}{1\linewidth}
    \includegraphics[width=1\linewidth]{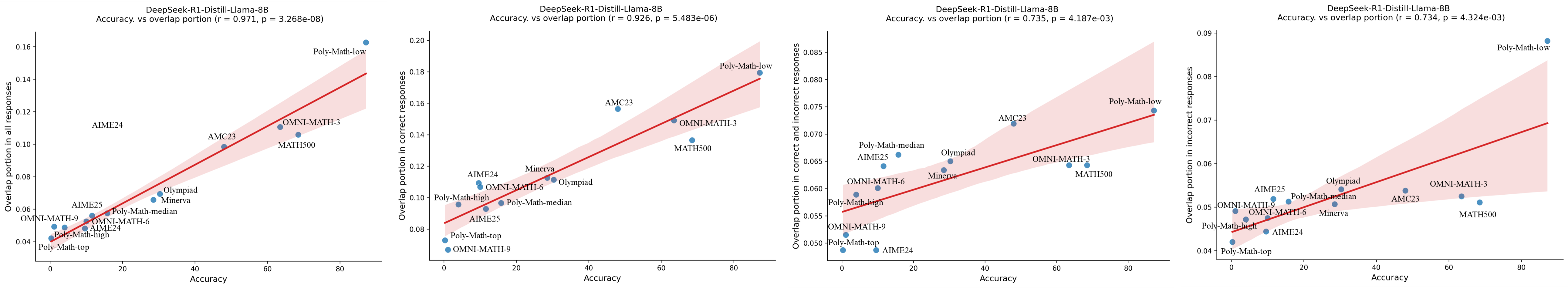}
    \subcaption{DeepSeek-R1-Distill-Llama-8B}
    \label{sub:ds-llama-8b}
  \end{subfigure}
  \\[5pt]
  
  \begin{subfigure}{1\linewidth}
    \includegraphics[width=1\linewidth]{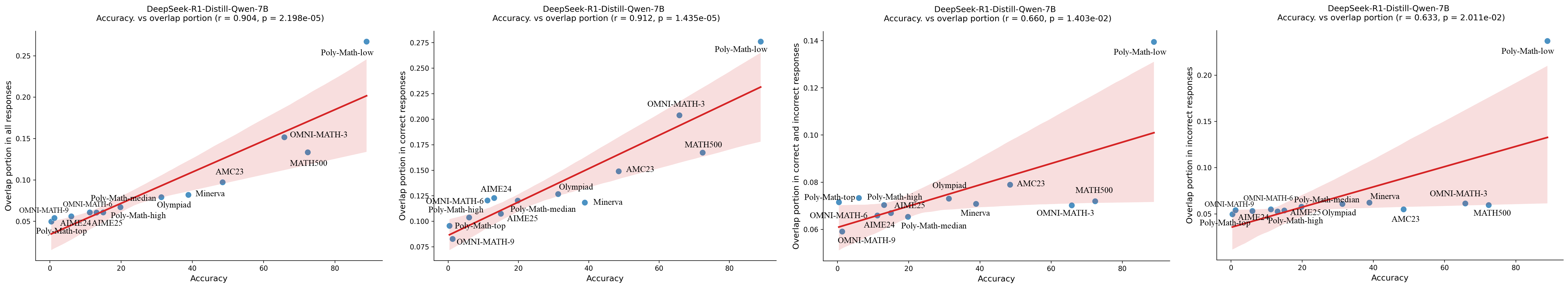}
    \subcaption{DeepSeek-R1-Distill-Qwen-7B}
    \label{sub:ds-qwen7b}
  \end{subfigure}
  \\[5pt]
  
  \begin{subfigure}{1\linewidth}
    \includegraphics[width=1\linewidth]{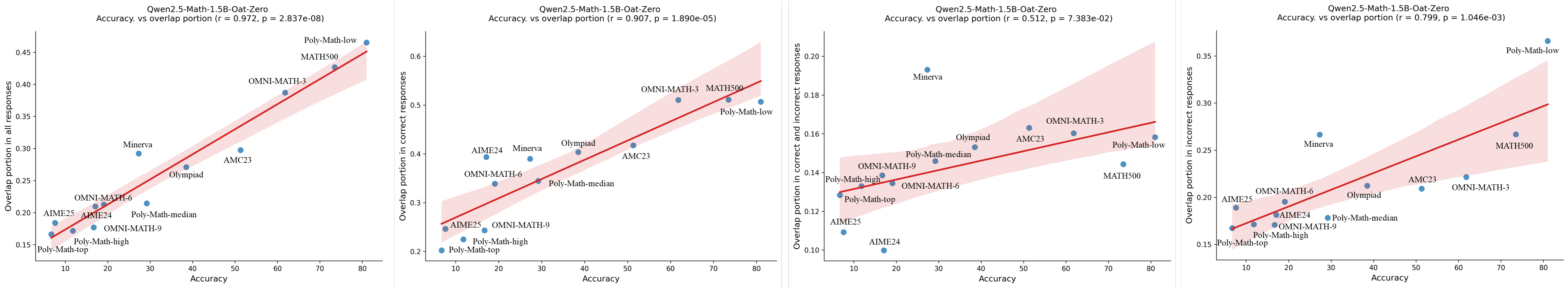}
    \subcaption{Qwen2.5-Math-1.5B-Oat-Zero}
    \label{sub:qwen2.5-1.5-math-zero}
  \end{subfigure}

  \caption{Additional correlations between accuracy and low-entropy segment overlap across backbones.
Panels (a)–(e) report, for Qwen2.5-Math-7B, Qwen2.5-7B-Base, DeepSeek-R1-Distill-Llama-8B, DeepSeek-R1-Distill-Qwen-7B, and Qwen2.5-Math-1.5B-Oat-Zero, the Pearson correlations between benchmark accuracy and four overlap ratios: all responses, correct-only responses, segments shared by correct and incorrect responses, and incorrect-only responses. Each point is a benchmark; the red line and shaded area show the fitted regression and its confidence band. Across models, accuracy is most strongly aligned with overlap among correct responses.}
  \label{fig:extra_corr}
\end{figure*}

Figure~\ref{fig:extra_corr} summarizes the results. For each backbone,
we compute four overlap ratios at the benchmark level:
(i) overlap among all responses, (ii) overlap among correct responses
only, (iii) overlap among segments shared by correct and incorrect
responses, and (iv) overlap among incorrect responses only.
We then correlate each ratio with benchmark accuracy.

Across all five backbones, we observe a consistent pattern: overlap among correct responses shows the strongest positive correlation with accuracy, overlap among all responses and shared segments yields weaker but still positive correlations, and overlap among incorrect-only segments is weakly correlated or even negatively correlated. These additional results support our claim that stable low-entropy structure in correct trajectories is a reliable indicator of reasoning quality, while overlap driven by incorrect trajectories does not translate into better performance.

\end{document}